%% file: eccv2020submission.tex
\documentclass[runningheads]{llncs}
\usepackage{graphicx}
\usepackage{comment}
\usepackage{amsmath,amssymb} 
\usepackage{color}
\usepackage{multirow}

\begin{document}
\pagestyle{headings}
\mainmatter
\def\ECCVSubNumber{4377}  

\title{Bridging Knowledge Graphs \\ to Generate Scene Graphs} 

\titlerunning{Bridging Knowledge Graphs to Generate Scene Graphs}
%
\author{Alireza Zareian \and
Svebor Karaman \and
Shih-Fu Chang}
\authorrunning{Alireza Zareian et al.}
%
\institute{Columbia University, New York NY 10027, USA \\
\email{\{az2407,sk4089,sc250\}@columbia.edu}}
\maketitle

\input{0_abstract}
\input{1_introduction}

\input{2_related_work}

\input{3_method}

\input{4_experiments}

\input{5_conclusion}

\clearpage

\bibliographystyle{splncs04}
\bibliography{egbib}

\clearpage

\appendix

\title{Supplementary Material}
\titlerunning{Bridging Knowledge Graphs to Generate Scene Graphs}
\author{}
\authorrunning{Alireza Zareian et al.}
\institute{}
\maketitle

\input{supp/1_introduction}
\input{supp/2_visualization}
\input{supp/3_time}

\input{supp/4_graph}

\input{supp/6_qualitative}

\input{supp/5_code}

\end{document}

%% file: 0_abstract.tex
\begin{abstract}
Scene graphs are powerful representations that parse images into their abstract semantic elements, \textit{i.e.}, objects and their interactions, which facilitates visual comprehension and explainable reasoning. On the other hand, commonsense knowledge graphs are rich repositories that encode how the world is structured, and how general concepts interact. 
In this paper, we present a unified formulation of these two constructs, where a scene graph is seen as an image-conditioned instantiation of a commonsense knowledge graph.
Based on this new perspective, we re-formulate scene graph generation as the inference of a bridge between the scene and commonsense graphs, where each entity or predicate instance in the scene graph has to be linked to its corresponding entity or predicate class in the commonsense graph. 
To this end, we propose a novel graph-based neural network that iteratively propagates information between the two graphs, as well as within each of them, while gradually refining their bridge in each iteration.
Our Graph Bridging Network, \textsc{GB-Net}, successively infers edges and nodes, allowing to simultaneously exploit and refine the rich, heterogeneous structure of the interconnected scene and commonsense graphs.
Through extensive experimentation, we showcase the superior accuracy of \textsc{GB-Net} compared to the most recent methods, 
resulting in a new state of the art.
We publicly release the source code of our method.\footnote{\url{https://github.com/alirezazareian/gbnet}}
\end{abstract}

%% file: 1_introduction.tex
\section{Introduction\label{sec:intro}}

\begin{figure}[t]
\begin{center}
\includegraphics[width=0.5\linewidth]{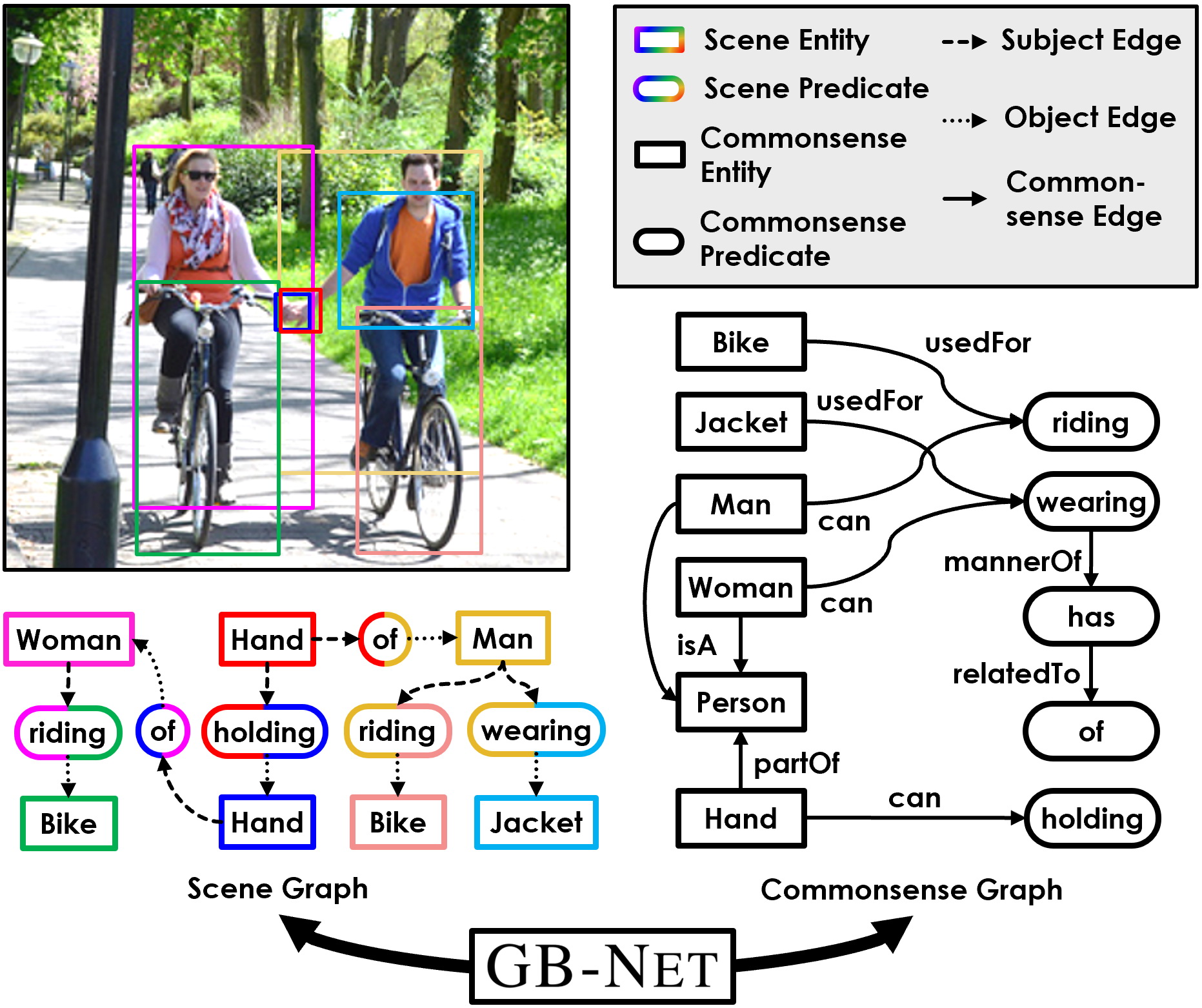}
\end{center}
   \caption{Left: An example of a Visual Genome image and its ground truth scene graph. Right: A relevant portion of the commonsense graph. In this paper we formulate the task of Scene Graph Generation as the problem of creating a bridge between these two graphs. Such bridge not only classifies each scene entity and predicate, but also creates an inter-connected heterogeneous graph whose rich structure is exploited by our method (\textsc{GB-Net}).}
\label{fig:vg_example}
\end{figure}

Extracting structured, symbolic, semantic representations from data has a long history in Natural Language Processing (NLP), under the umbrella terms \emph{semantic parsing} at the sentence level~\cite{gardner2018neural,flanigan2014discriminative} and \emph{information extraction} at the document level~\cite{li2019multilingual,wadden2019entity}. 
The resulting \emph{semantic graphs} or \emph{knowledge graphs} have many applications such as question answering~\cite{fader2014open,khashabi2018question} and information retrieval~\cite{dietz2018utilizing,yu2018modeling}.
In computer vision, Xu~\textit{et al.} have recently called attention to the task of Scene Graph Generation (SGG)~\cite{xu2017scene}, which aims at extracting a symbolic, graphical representation from a given image, where every node corresponds to a localized and categorized object (entity), and every edge encodes a pairwise interaction (predicate). 
This has inspired two lines of follow-up work, some improving the performance on SGG~\cite{li2017scene,newell2017pixels,zellers2018neural,yang2018graph,li2018factorizable,woo2018linknet,herzig2018mapping,gu2019scene,chen2019knowledge}, and others exploiting such rich structures for down-stream tasks such as Visual Question Answering (VQA)~\cite{teney2017graph,shi2019explainable,hudson2019learning,zhang2019empirical}, image captioning~\cite{yao2018exploring,yang2019auto}, image retrieval~\cite{johnson2015image,schuster2015generating}, and image synthesis \cite{johnson2018image}. 
In VQA for instance, SGG not only improves performance, but also promotes interpretability and enables explainable reasoning \cite{shi2019explainable}. 

Although several methods have been proposed, the state-of-the-art performance for SGG is still far from acceptable. 
Most recently, \cite{chen2019knowledge} achieves only 16\% mean recall, for matching the top 100 predicted subject-predicate-object triples against ground truth triples.
This suggests the current SGG methods are insufficient to address the complexity of this task.
Recently, a few papers have attempted to use external \emph{commonsense} knowledge to advance SGG~\cite{zellers2018neural,gu2019scene,chen2019knowledge}, as well as other domains \cite{chen2018iterative,kato2018compositional}. This commonsense can range from curated knowledge bases such as ConceptNet~\cite{liu2004conceptnet}, ontologies such as WordNet~\cite{miller1995wordnet}, or automatically extracted facts such as co-occurance frequencies \cite{zellers2018neural}.
The key message of those works is that a prior knowledge about the world can be very helpful when perceiving a complex scene. If we know the relationship of a \texttt{Person} and a \texttt{Bike} is most likely \texttt{riding}, we can more easily disambiguate between \texttt{riding}, \texttt{on}, and \texttt{attachedTo}, and classify their relationship more accurately. 
Similarly, if we know a \texttt{Man} and a \texttt{Woman} are both sub-types of \texttt{Person}, even if we only see \texttt{Man}-\texttt{riding}-\texttt{Bike} in training data, we can generalize and recognize a \texttt{Woman}-\texttt{riding}-\texttt{Bike} triplet at test time.
Although this idea is intuitively promising, existing methods that implement it have major limitations, as detailed in Section~\ref{sec:related_work}, and we address those in the proposed method. 

More specifically, recent methods either use ad-hoc heuristics to integrate limited types of commonsense into the scene graph generation process~\cite{chen2019knowledge,zellers2018neural}, or fail to exploit the rich, graphical structure of commonsense knowledge~\cite{gu2019scene}. To devise a general framework for incorporating any type of graphical knowledge into the process of scene understanding, we take inspiration from early works on knowledge representation and applying structured grammars to computer vision problems~\cite{zhao2011image,pei2011parsing,tu2014joint}, and redefine those concepts in the light of the recent advances in graph-based deep learning.
Simply put, we formulate both scene and commonsense graphs as knowledge graphs with entity and predicate nodes, and various types of edges.
A scene graph node represents an entity or predicate \emph{instance} in a specific image, while a commonsense graph node represents an entity or predicate \emph{class}, which is a general concept independent of the image. 
Similarly, a scene graph edge indicates the participation of an entity instance (\textit{e.g.} as a subject or object) in a predicate instance in a scene, while a commonsense edge states a general fact about the interaction of two concepts in the world. 
Figure~\ref{fig:vg_example} shows an example scene graph and commonsense graph side by side.

Based on this unified perspective, we reformulate the problem of scene graph generation from entity and predicate classification into the novel problem 
of bridging those two graphs. 
More specifically, we propose a method that given an image, initializes potential entity and predicate nodes, and then classifies each node by connecting it to its corresponding class node in the commonsense graph, through an edge we call a \textit{bridge}. 
This establishes a connectivity between instance-level, visual knowledge and generic, commonsense knowledge.
To incorporate the rich combination of visual and commonsense information in the SGG process, we propose a novel graphical neural network, that iteratively propagates messages between the scene and commonsense graphs, as well as within each of them, while gradually refining the bridge in each iteration. 
Our Graph Bridging Network, \textsc{GB-Net}, successively infers edges and nodes, allowing to simultaneously exploit and refine the rich, heterogeneous structure of the interconnected scene and commonsense graphs.


To evaluate the effectiveness of our method, we conduct extensive experiments on the Visual Genome~\cite{krishna2017visual} dataset. 
The proposed \textsc{GB-Net} outperforms the state of the art consistently in various performance metrics.
Through ablative studies, we show how each of the proposed ideas contribute to the results. 
We also publicly release a comprehensive software package based on~\cite{zellers2018neural} and~\cite{chen2019knowledge}, 
to reproduce all the numbers reported in this paper. We provide further quantitative, qualitative, and speed analysis in our Supplementary Material, as well as additional implementation details.

%% file: 2_related_work.tex
\section{Related work\label{sec:related_work}}

\subsection{Scene graph generation}

Most SGG methods are based on an object detection backbone that extracts region proposals from the input image. They utilize some kind of information propagation module to incorporate context, and then classify each region to an object class, as well as each pair of regions to a relation class~\cite{xu2017scene,zellers2018neural,yang2018graph,chen2019knowledge}. 
Our method has two key differences with this conventional process: firstly, our information propagation network operates on a larger graph which consists of not only object nodes, but also predicate nodes and commonsense graph nodes, and has a more complex structure. 
Secondly, we do not classify each object and relation using classifiers, but instead use a pairwise matching mechanism to connect them to corresponding class nodes in the commonsense graph.

More recently, a few methods~\cite{zellers2018neural,gu2019scene,chen2019knowledge} have used external knowledge to enhance scene graph generation.
This external knowledge is sometimes referred to as ``commonsense'', because it encodes ontological knowledge about classes, rather than specific instances. Despite encouraging results, these methods have major limitations. 
Specifically, \cite{zellers2018neural} used triplet frequency to bias the logits of their predicate classifier, and \cite{chen2019knowledge} used such frequencies to initialize edge weights 
on their graphs. 
Such external priors have been also shown beneficial for recognizing objects~\cite{xu2019spatial,xu2019reasoning} and relationships~\cite{liang2018visual,zhan2019exploring}, that are building blocks for SGG. 
Nevertheless, neither of those methods can incorporate other types or knowledge, such as the semantic hierarchy concepts, or object affordances. 
Gu~\textit{et al.}~\cite{gu2019scene} propose a more general way to incorporate knowledge in SGG, by retrieving a set of relevant facts for each object from a pool of commonsense facts. 
However, their method does not utilize the structure of the commonsense graph, and treats knowledge as a set of triplets. 
Our method considers commonsense as a general graph with several types of edges, explicitly integrates that graph with the scene graph by connecting corresponding nodes, and incorporates the rich structure of commonsense by graphical message passing.

\subsection{Graph-based neural networks}

By Graph-based Neural Networks (GNN), we refer to the family of models that take a graph as input, and iteratively update the representation of each node by applying a learnable function (a.k.a., message) on the node's neighbors. Graph Convolutional Networks (GCN) \cite{kipf2016semi}, Gated Graph Neural Networks (GGNN) \cite{li2015gated}, and others are all specific implementations of this general model. Most SGG methods use some variant of GNNs to propagate information between region proposals~\cite{xu2017scene,li2017scene,yang2018graph,chen2019knowledge}. 
Our message passing method, detailed in Section~\ref{sec:method}, resembles GGNN  but instead of propagating messages through a static graph, we update (some) edges as well. Few methods exist that dynamically update edges during message passing~\cite{qi2018learning,zareian2020weakly}, but we are the first to refine edges between a scene graph and an external knowledge graph.

Apart from SGG, GNNs have been used in several other computer vision tasks, often in order to propagate context information across different objects in a scene. For instance, \cite{liu2018structure} injects a GNN into a Faster R-CNN \cite{ren2015faster} framework to contextualize the features of region proposals before classifying them. This improves the results since the presence of a \texttt{table} can affect the detection of a \texttt{chair}. On the other hand, some methods utilize GNNs on graphs that represent the ontology of concepts, rather than objects in a scene~\cite{marino2016more,wang2018zero,lee2018multi,kato2018compositional}. This often enables generalization to unseen or infrequent concepts by incorporating their relationship with frequently seen concepts.
More similarly to our work, Chen~\textit{et al.}~\cite{chen2018iterative} were the first to bring those two ideas together, and form a graph by objects in an image as well as object classes in the ontology. Nevertheless, the class nodes in that work were merely an auxiliary means to improve object features before classification. 
In contrast, we classify the nodes by explicitly inferring their connection to their corresponding class nodes. Moreover, we iteratively refine the bridge between scene and commonsense graphs to enhance our prediction. Furthermore, their task only involves objects and object classes, while we explore a more complex structure where predicates play an important role as well.

%% file: 3_method.tex
\section{Problem Formulation\label{sec:formulation}}

In this section, we first formalize the concepts of knowledge graph in general, and commonsense graph and scene graph in particular. 
Leveraging their similarities, we then reformulate the problem of scene graph generation as bridging these two graphs. 


\subsection{Knowledge graphs}

We define a knowledge graph as a set of entity and predicate nodes $(\mathcal{N}_\text{E},\mathcal{N}_\text{P})$, each with a semantic label, and a set of directed, weighted edges $\mathcal{E}$ from a predefined set of types. 
Denoting by $\Delta$ a node type (here, either entity E or predicate P), the set of edges encoding the relation $r$ between nodes of type $\Delta$ and $\Delta'$ is defined as
\begin{equation}
\begin{aligned}
\mathcal{E}^{\Delta\rightarrow\Delta'}_r \subseteq \mathcal{N}_\Delta \times \mathcal{N}_{\Delta'} \rightarrow \mathbb{R}.
\end{aligned}
\end{equation}

\paragraph{A commonsense graph} is a type of knowledge graph in which each node represents the general concept of its semantic label, and hence each semantic label (entity or predicate class) appears in exactly one node. 
In such a graph, each edge encodes a relational fact involving a pair of concepts, such as \texttt{Hand-partOf-Person} and \texttt{Cup-usedFor-Drinking}. 
Formally, we define the set of commonsense entity (CE) nodes $\mathcal{N}_\text{CE}$ and commonsense predicate (CP) nodes $\mathcal{N}_\text{CP}$ as all entity and predicate classes in our task. 
Commonsense edges $\mathcal{E}_\text{C}$ consist of 4 distinct subsets, depending on the source and destination node type:
\begin{equation}
\begin{aligned}
\mathcal{E}_\text{C} = &\{\mathcal{E}^{\text{CE}\rightarrow\text{CP}}_r\} \cup \{\mathcal{E}^{\text{CP}\rightarrow\text{CE}}_r\} \, \cup
\\ &
\{\mathcal{E}^{\text{CE}\rightarrow\text{CE}}_r\} \cup \{\mathcal{E}^{\text{CP}\rightarrow\text{CP}}_r\}.
\end{aligned}
\end{equation}


\paragraph{A scene graph} is a different type of knowledge graph where: (a) each scene entity (SE) node is associated with a bounding box, referring to an image region, (b) each scene predicate (SP) node is associated with an ordered pair of SE nodes, namely a subject and an object, and (c) there are two types of undirected edges which connect each SP to its corresponding subject and object respectively. Here because we define knowledge edges to be directed, we model each undirected subject or object edge as two directed edges in the opposite directions, each with a distinct type. More specifically, 
\begin{equation}
\begin{aligned}
    \mathcal{N}_\text{SE} \subseteq& [0,1]^4 \times \mathcal{N}_\text{CE}
    ,\\
    \mathcal{N}_\text{SP} \subseteq& \mathcal{N}_\text{SE} \times \mathcal{N}_\text{SE} \times \mathcal{N}_\text{CP}
    ,\\
    \mathcal{E}_\text{S} =& \{\mathcal{E}^{\text{SE}\rightarrow\text{SP}}_\texttt{subjectOf}, \mathcal{E}^{\text{SE}\rightarrow\text{SP}}_\texttt{objectOf},\\&\;\, \mathcal{E}^{\text{SP}\rightarrow\text{SE}}_\texttt{hasSubject}, \mathcal{E}^{\text{SP}\rightarrow\text{SE}}_\texttt{hasObject}\}
,\end{aligned}
\end{equation}
where $[0,1]^4$ is the set of possible bounding boxes, and $\mathcal{N}_\text{SE} \times \mathcal{N}_\text{SE} \times \mathcal{N}_\text{CP}$ is the set of all possible triples that consist of two scene entity nodes and a scene predicate node.
Figure~\ref{fig:vg_example} shows an example of scene graph and commonsense graph side by side, to make their similarities clearer.
Here we assume every scene graph node has a label that exists in the commonsense graph, since in reality some objects and predicates might belong to background classes, we consider a special commonsense node as background entity and another for background predicate. 

\subsection{Bridging knowledge graphs}

Considering the similarity between the commonsense and scene graph  formulations, we make a subtle refinement in the formulation to bridge these two graphs. 
Specifically, we remove the class from SE and SP nodes and instead encode it into a set of \emph{bridge} edges $\mathcal{E}_\text{B}$ that connect each SE or SP node to its corresponding class, \textit{i.e.}, a CE or CP node respectively:
\begin{equation}
\begin{aligned}
    \mathcal{N}_\text{SE}^\textbf{ ?} \subseteq & \, [0,1]^4 
    ,\\
    \mathcal{N}_\text{SP}^\textbf{ ?} \subseteq & \, \mathcal{N}_\text{SE} \times \mathcal{N}_\text{SE}
    ,\\
    \mathcal{E}_\text{B} = & \, \{\mathcal{E}^{\text{SE}\rightarrow\text{CE}}_\texttt{classifiedTo}, \mathcal{E}^{\text{SP}\rightarrow\text{CP}}_\texttt{classifiedTo},\\&\;\, \mathcal{E}^{\text{CE}\rightarrow\text{SE}}_\texttt{hasInstance}, \mathcal{E}^{\text{CP}\rightarrow\text{SP}}_\texttt{hasInstance}\}
,\end{aligned}
\end{equation}
where $.^\textbf{?}$ means the nodes are implicit, \textit{i.e.}, their classes are unknown. 
Each edge of type \texttt{classifiedTo}, connects an entity or predicate to its corresponding label in the commonsense graph, and has a reverse edge of type \texttt{hasInstance} which connects the commonsense node back to the instance. 
Based on this reformulation, we can define the problem of SGG as the extraction of implicit entity and predicate nodes from the image (hereafter called \textit{scene graph proposal}), and then classifying them by connecting each entity or predicate to the corresponding node in the commonsense graph. 
Accordingly, 
Given an input image $I$ and a provided and fixed commonsense graph, the goal of SGG with commonsense knowledge is to maximize
\begin{equation}
\label{eq:prob}
\begin{aligned}
&p(\mathcal{N}_\text{SE}, \mathcal{N}_\text{SP}, \mathcal{E}_\text{S} | I, \mathcal{N}_\text{CE}, \mathcal{N}_\text{CP}, \mathcal{E}_\text{C}) = \\ &\qquad p(\mathcal{N}_\text{SE}^\textbf{ ?}, \mathcal{N}_\text{SP}^\textbf{ ?}, \mathcal{E}_\text{S} | I) \times \\
&\qquad p(\mathcal{E}_\text{B} | I, \mathcal{N}_\text{CE}, \mathcal{N}_\text{CP}, \mathcal{E}_\text{C}, \mathcal{N}_\text{SE}^\textbf{ ?}, \mathcal{N}_\text{SP}^\textbf{ ?}, \mathcal{E}_\text{S}) 
.\end{aligned}
\end{equation}

In this paper, the first term is implemented as a region proposal network that infers $\mathcal{N}_\text{SE}^\textbf{ ?}$ given the image, followed by a simple predicate proposal algorithm that considers all possible entity pairs as $\mathcal{N}_\text{SP}^\textbf{ ?}$.
The second term is fulfilled by the proposed \textsc{GB-Net} which infers bridge edges by incorporating the rich structure of the scene and commonsense graphs. 
Note that unlike most existing methods \cite{zellers2018neural,chen2019knowledge}, we do not factorize this into predicting entity classes given the image, and then predicate classes given entities. Therefore, our formulation is more general and allows the proposed method to classify entities and predicates jointly. 



\begin{figure}[t]
\begin{center}
\includegraphics[width=1.0\linewidth]{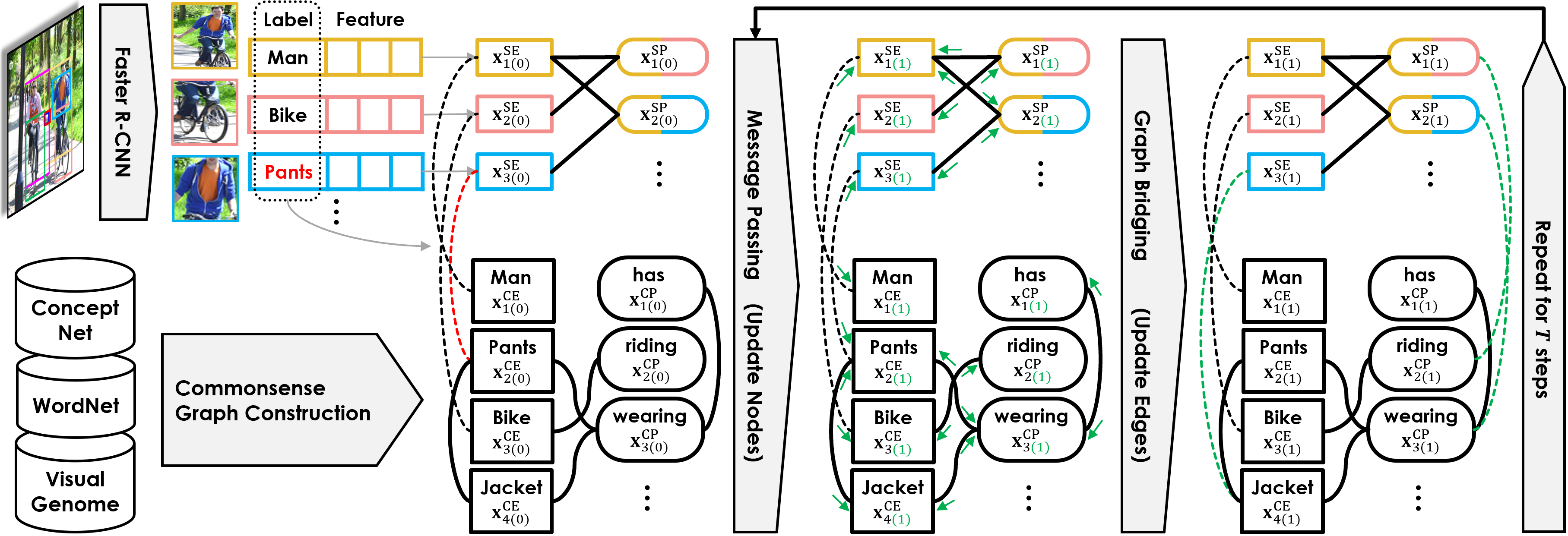}
\end{center}
   \caption{An illustrative example of the \textsc{GB-Net} process. First, we initialize the scene graph and entity bridges using a Faster R-CNN. Then we propagate messages to update node representations, and use them to update the entity and predicate bridges. This is repeated $T$ times and the final bridge determines the output label of each node.}
\label{fig:overview}
\end{figure}

\section{Method\label{sec:method}}


The proposed method is illustrated in Figure~\ref{fig:overview}. 
Given an image, our model first applies a Faster R-CNN~\cite{ren2015faster} to detect objects, and represents them as scene entity (SE) nodes. 
It also creates a scene predicate (SP) node for each pair of entities, which forms a scene graph proposal, yet to be classified. 
Given this graph and a background commonsense graph, each with fixed internal connectivity, our goal is to create \emph{bridge} edges between the two graphs that connect each instance (SE and SP node) to its corresponding class (CE and CP node). 
To this end, our model initializes entity bridges by connecting each SE to the CE that matches the label predicted by Faster R-CNN, and propagates messages among all nodes, through every edge type with dedicated message passing parameters. 
Given the updated node representations, it computes a pairwise similarity between every SP node and every CP node, and finds maximal similarity pairs to connect scene predicates to their corresponding classes, via predicate bridges.
It also does the same for entity nodes to potentially refine their bridges too.
Given the new bridges, it propagates messages again, and repeats this process for a predefined number of steps. 
The final state of the bridge determines which class each node belongs to, resulting in the output scene graph. 

\subsection{Graph initialization}

The object detector outputs a set of $n$ detected objects, each with a bounding box $b_j$, a label distribution $p_j$ and an RoI-aligned~\cite{ren2015faster} feature vector $\mathbf{v}_j$. Then we allocate a \emph{scene entity node} (SE) for each object, and a \emph{scene predicate node} (SP) for each pair of objects, representing the potential predicate with the two entities as its subject and object.
Each entity is initialized using its RoI features $\mathbf{v}_j$, and each predicate is initialized using the RoI features $\mathbf{u}_j$ of a bounding box enclosing the union of its subject and object.
Formally, we can write,
\textit{i.e.},
\begin{equation}
\begin{aligned}
\mathbf{x}^{\text{SE}}_j = \phi^{\text{SE}}_\text{init}(\mathbf{v}_j)
\;, \quad \text{and} \quad
\mathbf{x}^{\text{SP}}_j = \phi^{\text{SP}}_\text{init}(\mathbf{u}_j)
,\end{aligned}
\end{equation}
where $\phi^\text{SE}_\text{init}$ and $\phi^\text{SP}_\text{init}$ are two fully connected networks that are branched from the backbone after ROI-align.
To form a scene graph proposal, we connect each predicate node to its subject and object via labeled edges. Specifically, we define the following 4 edge types: for a triplet $s-p-o$, we connect $p$ to $s$ using a \texttt{hasSubject} edge, $p$ to $o$ using a \texttt{hasObject} edge, $s$ to $p$ using a \texttt{subjectOf} edge, and $o$ to $p$ using an \texttt{objectOf} edge. 
The reason we have two directions as separate types is that in the message passing phase, the way we use predicate information to update entities should be different from the way we use entities to update predicates. 

On the other hand, we initialize the commonsense graph with \emph{commonsense entity nodes} (CE) and \emph{commonsense predicate nodes} (CP) using a linear projection of their word embeddings:
\begin{equation}
\begin{aligned}
&\mathbf{x}^{\text{CE}}_i = \phi^{\text{CE}}_\text{init}(\mathbf{e}^n_i)
\;, \quad \text{and} \quad
&\mathbf{x}^{\text{CP}}_i = \phi^{\text{CP}}_\text{init}(\mathbf{e}^p_i)
.\end{aligned}
\end{equation}
The commonsense graph also has various types of edges, such as \texttt{UsedFor} and \texttt{PartOf}, as detailed in Section~\ref{sec:implementation_details}. Our method is independent of the types of commonsense edges, and can utilize any provided graph from any source.

So far, we have two isolated graphs, scene and commonsense. 
An SE node representing a detected \texttt{Person} intuitively refers to the \texttt{Person} concept in the ontology, and hence the \texttt{Person} node in the commonsense graph. 
Therefore, we connect each SE node to the CE node that corresponds the semantic label predicted by Faster R-CNN, via a 
$\texttt{classifiedTo}$
edge type. Instead of a hard classification, we connect each entity to top $K_\text{bridge}$ classes using $p_j$ (class distribution predicted by Faster R-CNN) as weights. 
We also create a reverse connection from each CE node to corresponding SE nodes, 
using an 
$\texttt{hasInstance}$ edge, 
but with the same weights $p_j$. 
As mentioned earlier, this is to make sure information flows from commonsense to scene as well as scene to commonsense, but not in the same way.
We similarly define two other edge types, 
$\texttt{classifiedTo}$ and $\texttt{hasInstance}$ 
for predicates, which are initially an empty set, and will be updated to bridge SP nodes to CP nodes as we explain in the following. 
These 4 edge types can be seen as flexible \emph{bridges} that connect the two fixed graphs, which are considered latent variables to be determined by the model. 

This forms a heterogeneous graph with four types of nodes (SE, SP, CE, and CP) and various types of edges: scene graph edges $\mathcal{E}_\text{S}$ such as \texttt{subjectOf}, commonsense edges $\mathcal{E}_\text{C}$ such as \texttt{usedFor}, and bridge edges $\mathcal{E}_\text{B}$ such as \texttt{classifiedTo}. 
Next, we explain how our proposed method updates node representations and bridge edges, while keeps commonsense and scene edges constant.

\subsection{Successive message passing and bridging}

Given a heterogeneous graph as described above, we employ a variant of GGNN~\cite{li2015gated} to propagate information among nodes. First, each node representation is fed into a fully connected network to compute \emph{outgoing} messages, that is 
\begin{equation}
\begin{aligned}
\mathbf{m}^{\Delta\rightarrow}_i = \phi^{\Delta}_\text{send}(\mathbf{x}^{\Delta}_i)
,\end{aligned}
\end{equation}
for each $i$ and node type $\Delta$, where $\phi_\text{send}$ is a trainable \emph{send head} which has shared weights across nodes of each type. 
After computing outgoing messages, we 
send them through all outgoing edges, multiplying by the edge weight. Then for each node, we aggregate incoming messages, by first adding across edges of the same type, and then concatenating across edge types. We compute the \emph{incoming} message for each node by applying another fully connected network on the aggregated messages:
\begin{equation}
\begin{aligned}
\mathbf{m}^{\Delta\leftarrow}_j = \phi^{\Delta}_\text{receive}\Bigg(\bigcup_{\Delta'}\bigcup^{\mathcal{E}_k \in \mathcal{E}^{\Delta' \rightarrow \Delta}}\sum_{(i,j, a^k_{ij}) \in \mathcal{E}_k} a^k_{ij} \mathbf{m}^{\Delta'\rightarrow}_i\Bigg)
,\end{aligned}
\end{equation}
where $\phi_\text{receive}$ is a trainable \emph{receive head} and $\cup$ denotes concatenation. 
Note that the first concatenation is over all 4 node types, the second concatenation is over all edge types from $\Delta'$ to $\Delta$, and the sum is over all edges of that type, where $i$ and $j$ are the head and tail nodes, and $a^k_{ij}$ is the edge weight.
Given the incoming message for each node, we update the representation of the node using a Gated Recurrent Unit (GRU) update rule, following \cite{cho2014learning}:
\begin{equation}
\begin{aligned}
\mathbf{z}_j^\Delta &= \sigma\big(W_z^\Delta \mathbf{m}^{\Delta\leftarrow}_j + U_z^\Delta \mathbf{x}_j^\Delta\big)
,\\
\mathbf{r}_j^\Delta &= \sigma\big(W_r^\Delta \mathbf{m}^{\Delta\leftarrow}_j + U_r^\Delta \mathbf{x}_j^\Delta\big)
,\\
\mathbf{h}_j^\Delta &= \tanh\big(W_h^\Delta \mathbf{m}^{\Delta\leftarrow}_j + U_h^\Delta (\mathbf{r}_j^\Delta \odot \mathbf{x}_j^\Delta)\big)
,\\
\mathbf{x}_j^\Delta &\Leftarrow (1-\mathbf{z}_j^\Delta)\odot \mathbf{x}_j^\Delta + \mathbf{z}_j^\Delta\odot \mathbf{h}_j^\Delta
,\end{aligned}
\end{equation}
where $\sigma$ is the sigmoid function, and $W_.^\Delta$ and $U_.^\Delta$ are trainable matrices that are shared across nodes of the same type, but distinct for each node type $\Delta$. This update rule can be seen as an extension of GGNN \cite{li2015gated} to heterogeneous graphs, with a more complex message aggregation strategy. Note that $\Leftarrow$ means we update the node representation. Mathematically, this means $\mathbf{x}_{j(t+1)}^\Delta = U(\mathbf{x}_{j(t)}^\Delta)$, where $U$ is the aforementioned update rule and $(t)$ denotes iteration number. For simplicity, we drop this subscript throughout this paper.

So far, we have explained how to update node representations using graph edges. 
Now using the new node representations, we should update the bridge edges $\mathcal{E}_\text{B}$
that connect scene nodes to commonsense nodes. 
To this end, we compute a pairwise similarity from each SE to all CE nodes, and from each SP to all CP nodes. 
\begin{equation}
\begin{aligned}
\mathbf{a}^\text{EB}_{ij} = \frac{\exp \langle \mathbf{x}_i^\text{SE}, \mathbf{x}_j^\text{CE} \rangle_\text{EB}}{\sum_{j'} \exp \langle \mathbf{x}_i^\text{SE}, \mathbf{x}_{j'}^\text{CE} \rangle_\text{EB}} 
\;, \quad \text{where} \quad
\langle \mathbf{x}, \mathbf{y} \rangle_\text{EB} = \phi^{\text{SE}}_\text{att}(\mathbf{x})^T \phi^{\text{CE}}_\text{att}(\mathbf{y})
,\end{aligned}
\end{equation}
and similarly for predicates,
\begin{equation}
\begin{aligned}
\mathbf{a}^\text{PB}_{ij} = \frac{\exp \langle \mathbf{x}_i^\text{SP}, \mathbf{x}_j^\text{CP} \rangle_\text{PB}}{\sum_{j'} \exp \langle \mathbf{x}_i^\text{SP}, \mathbf{x}_{j'}^\text{CP} \rangle_\text{PB}} 
\;, \quad \text{where} \quad
\langle \mathbf{x}, \mathbf{y} \rangle_\text{PB} = \phi^{\text{SP}}_\text{att}(\mathbf{x})^T \phi^{\text{CP}}_\text{att}(\mathbf{y})
.\end{aligned}
\end{equation}
Here $\phi^{\Delta}_\text{att}$ is a fully connected network that resembles \emph{attention head} in transformers. Note that since $\phi^{\Delta}_\text{att}$ is not shared across node types, our similarity metric is asymmetric. 
We use each $\mathbf{a}^{\text{EB}}_{ij}$ to set the edge weight of the 
\texttt{classifiedTo} 
edge from $\mathbf{x}_i^\text{SE}$ to $\mathbf{x}_j^\text{CE}$, as well as the 
\texttt{hasInstance} 
edge from $\mathbf{x}_j^\text{CE}$ to $\mathbf{x}_i^\text{SE}$. 
Similarly we use each $\mathbf{a}^{\text{PB}}_{ij}$ to
set the weight of edges between $\mathbf{x}_i^\text{SP}$ and $\mathbf{x}_j^\text{CP}$.
In preliminary experiments we realised that such fully connected bridges hurt performance in large graphs. 
Hence, we only keep the top $K_\text{bridge}$ values of $\mathbf{a}^{\text{EB}}_{ij}$ for each $i$, and set the rest to zero. We do the same thing for predicates, keeping the top $K_\text{bridge}$ values of $\mathbf{a}^{\text{PB}}_{ij}$ for each $i$.
Given the updated bridges, we propagate messages again to update node representations, and iterate for a fixed number of steps, $T$. 
The final values of $\mathbf{a}^{\text{EB}}_{ij}$ and $\mathbf{a}^{\text{PB}}_{ij}$ are the outputs of our model, which can be used to classify each entity and predicate in the scene graph. 

\subsection{Training}
We closely follow \cite{chen2019knowledge} which itself follows \cite{zellers2018neural} for training procedure. Specifically, given the output and ground truth graphs, we align output entities and predicates to ground truth counterparts. To align entities we use IoU and predicates will be aligned naturally since they correspond to aligned pairs of entities. Then we use the output probability scores of each node to define a cross-entropy loss. The sum of all node-level loss values will be the objective function to be minimized using Adam \cite{kingma2014adam}.

Due to the highly imbalanced predicate statistics in Visual Genome, we observed that best-performing models usually concentrate their performance merely on the most frequent classes such as \texttt{on} and \texttt{wearing}. To alleviate this, we modify the basic cross-entropy objective that is commonly used by assigning an importance weight to each class. We follow the recently proposed class-balanced loss \cite{cui2019class} where the weight of each class is inversely proportional to its frequency. More specifically, we use the following loss function for each predicate node:
\begin{equation}
\begin{aligned}
\mathcal{L}^P_i = - \frac{1 - \beta}{1 - \beta^{n_j}} \log \mathbf{a}^{\text{PB}}_{ij}
,\end{aligned}
\end{equation}
where $j$ is the class index of the ground truth predicate aligned with $i$, $n_j$ is the frequency of class $j$ in training data, and $\beta$ is a hyperparameter. Note that $\beta=0$ leads to a regular cross-entropy loss, and the more it approaches $1$, the more strictly it suppresses frequent classes.
To be fair in comparison with other methods, we include a variant of our method without reweighting, which still outperforms all other methods.


%% file: 4_experiments.tex
\section{Experiments\label{sec:experiments}}

Following the literature, we use the large-scale Visual Genome benchmark \cite{krishna2017visual} to evaluate our method. 
We first show our \textsc{GB-Net} outperforms the state of the art, by extensively evaluating it on 24 performance metrics.
Then we present an ablation study to illustrate how each innovation contributes to the performance. In the Supplementary Material, we also provide a per-class performance breakdown to show the consistency and robustness of our performance across frequent and rare classes. That is accompanied by a computational speed analysis, and several qualitative examples of our generated graphs compared to the state of the art, side by side.

\subsection{Task description}

Visual Genome \cite{krishna2017visual} consists of 108,077 images with annotated objects (entities) and pairwise relationships (predicates), which is then post-processed by \cite{xu2017scene} to create scene graphs. They use the most frequent 150 entity classes and 50 predicate classes to filter the annotations. Figure~\ref{fig:vg_example} shows an example of their post-processed scene graphs which we use as ground truth. We closely follow their evaluation settings such as train and test splits. 

The task of scene graph generation, as described in Section~\ref{sec:method}, is equivalent to the \textsc{SGGen} scenario proposed by \cite{xu2017scene} and followed ever since. Given an image, the task of \textsc{SGGen} is to jointly infer entities and predicates from scratch. Since this task is limited by the quality of the object proposals, \cite{xu2017scene} also introduced two other tasks that more clearly evaluate entity and predicate recognition. In \textsc{SGCls}, we take localization (here region proposal network) out of the picture, by providing the model with ground truth bounding boxes during test, simulating a \textit{perfect} proposal model. 
In \textsc{PredCls}, we take object detection for granted, and provide the model with not only ground truth bounding boxes, but also their true entity class. 
In each task, the main evaluation metric is average per-image recall of the top K subject-predicate-object triplets. The confidence of a triplet that is used for ranking is computed by multiplying the classification confidence of all three elements. Given the ground truth scene graph, each predicate forms a triplet, which we match against the top $K$ triplets in the output scene graph. A triplet is matched if all three elements are classified correctly, and the bounding boxes of subject and object match with an IoU of at least $0.5$. 
Besides the choice of $K$, there are two other choices to be made: (1) Whether or not to enforce the so-called \emph{Graph Constraint} (GC), which limits the top K triplets to only one predicate for each ordered entity pair, and (2) Whether to compute the recall for each predicate class separately and take the mean (mR), or compute a single recall for all triplets (R)~\cite{chen2019knowledge}. 
We comprehensively report both mean and overall recall, both with and without GC, and conventionally use both 50 and 100 for $K$, resulting in 8 metrics for each task, 24 in total.

\subsection{Implementation details}
\label{sec:implementation_details}

We use three-layer fully connected networks with ReLU activation for all trainable networks $\phi_\text{init}$, $\phi_\text{send}$, $\phi_\text{receive}$ and $\phi_\text{att}$. We set the dimension of node representations to 1024, and perform 3 message passing steps, except in ablation experiments where we try 1, 2 and 3. We tried various values for $\beta$. Generally the higher it is, mean recall improves and recall falls. We found $0.999$ is a good trade-off, and chose $K_\text{bridge}=5$ empirically. All hyperparameters are tuned using a validation set randomly selected from training data. We borrow the Faster R-CNN trained by \cite{zellers2018neural} and shared among all our baselines, which has a VGG-16 backbone and predicts 128 proposals. 

In our commonsense graph, the nodes are the 151 entity classes and 51 predicate classes that are fixed by \cite{xu2017scene}, including background. We use the GloVE \cite{pennington2014glove} embedding of category titles to initialize their node representation (via $\phi_\text{init}$), and fix GloVE during training. 
We compile our commonsense edges from three sources, WordNet \cite{miller1995wordnet}, ConceptNet \cite{liu2004conceptnet}, and Visual Genome.
To summarize, there are three groups of edge types in our commonsense graph. We have \texttt{SimilarTo} from WordNet hierarchy, we have \texttt{PartOf}, \texttt{RelatedTo}, \texttt{IsA}, \texttt{MannerOf}, and \texttt{UsedFor} from ConceptNet, and finally from VG training data we have conditional probabilities of 
subject given predicate, predicate given subject, subject given object, \textit{etc.}
We explain the details in the supplementary material.
The process of compiling and pruning the knowledge graph is semi-automatic and takes less than a day from a single person. We make it publicly available as a part of our code. We have also tried using each individual source (e.g. only ConceptNet) independently, which requires less effort, and does not significantly impact the performance.
There are also recent approaches to automate the process of commonsense knowledge graph construction~\cite{bosselut2019comet,ilievski2020consolidating}, which can be utilized to further reduce the manual labor.

\subsection{Main results}
\input{tables/main}

Table~\ref{table:allresults} summarizes 
our results in comparison to the state of the art. IMP+ refers to the re-implementation of \cite{xu2017scene} by \cite{zellers2018neural} using their new Faster R-CNN backbone. That method does not use any external knowledge and only uses message passing among the entities and predicates and then classifies each. Hence, it can be seen as a strong, but knowledge-free baseline. FREQ is a simple baseline proposed by \cite{zellers2018neural}, which predicts the most frequent predicate for any given pair of entity classes, solely based on statistics from the training data.
FREQ surprisingly outperforms IMP+, confirming the efficacy of commonsense in SGG. 

SMN \cite{zellers2018neural} applies bi-directional LSTMs on top of the entity features, then classifies each entity and each pair. They bias their classifier logits using statistics from FREQ, which improves their total recall significantly, at the expense of higher bias against less frequent classes, as revealed by \cite{chen2019knowledge}. More recently, KERN \cite{chen2019knowledge} 
encodes VG statistics into the edge weights of the graph, which is then incorporated by propagating messages.
Since it encodes statistics more implicitly, KERN is less biased compared to SMN, which improves mR. Our method improves both R and mR significantly, and our class-balanced model, \textsc{GB-Net}-$\beta$, further enhances mR ($+2.7$\% in average) without hurting R by much ($-0.2$\%).

We observed that the state of the art performance has been saturated in the \textsc{SGGen} setting, especially for overall recall. 
This is partly because object detection performance is a bottleneck that limits the performance. It is worth noting that mean recall is a more important metric than overall recall, since most SGG methods tend to score a high overall recall by investing on few most frequent classes, and ignoring the rest~\cite{chen2019knowledge}. As shown in Table \ref{table:allresults}, our method achieves significant improvements in mean recall. We provide in-depth performance analysis by comparing our recall per predicate class with that of the state of the art, as well as qualitative analysis in the Supplementary Material.

There are other recent SGG methods that are not used for comparison here, because their evaluation settings are not identical to ours, and their code is not publicly available to the best of our knowledge~\cite{gu2019scene,qi2019attentive}. For instance, \cite{qi2019attentive} reports only 8 out of our 24 evaluation metrics, and although our method is superior in 6 metrics out of those 8, that is not sufficient to fairly compare the two methods. 

\subsection{Ablation study}

\input{tables/ablation}


To further explain our performance improvement, Table \ref{table:ablation} compares our full method with its weaker variants. Specifically, to investigate the effectiveness of commonsense knowledge, we remove the commonsense graph and instead classify each node in our graph using a 2-layer fully connected classifier after message passing. This negatively impacts performance in all metrics, proving our method is able to exploit commonsense knowledge through the proposed bridging technique.
Moreover, to highlight the importance of our proposed message passing and bridge refinement process, we repeated the experiments with fewer steps.
We observe the performance drops significantly with fewer steps, proving the effectiveness of our model, but it saturates as we go beyond 3 steps.

%% file: tables/main.tex
\begin{table}[t]
\begin{center}
\caption{\label{table:allresults}
Evaluation in terms of mean and overall triplet recall, at top 50 and top 100, with and without Graph Constraint (GC), for the three tasks of \textsc{SGGen}, \textsc{SGCls} and \textsc{PredCls}. Numbers are in percentage. All baseline numbers were borrowed from \cite{chen2019knowledge}. Top two methods for each metric is shown in \textbf{bold} and \textit{italic} respectively.}
\begin{tabular}{c|c|c| c c c c | c c}
\hline


\multirow{2}{*}{Task} & \multirow{2}{*}{Metric} & \multirow{2}{*}{GC} & \multicolumn{6}{c}{Method} \\
&&& IMP+ & FREQ & SMN & KERN & \textsc{GB-Net} & \textsc{GB-Net}-$\beta$ \\

\hline\hline
\multirow{8}{*}{\textsc{SGGen}} & \multirow{2}{*}{mR@50} & Y & 3.8 & 4.3 & 5.3 & \textit{6.4} & 6.1 & \textbf{7.1} \\
&& N & 5.4 & 5.9 & 9.3 & \textit{11.7} & 9.8 & \textbf{11.7} \\
& \multirow{2}{*}{mR@100} & Y & 4.8 & 5.6 & 6.1 & 7.3 & \textit{7.3} & \textbf{8.5} \\
&& N & 8.0 & 8.9 & 12.9 & \textit{16.0} & 14.0 & \textbf{16.6} \\
& \multirow{2}{*}{R@50} & Y & 20.7 & 23.5 & \textbf{27.2} & \textit{27.1} & 26.4 & 26.3 \\
&& N & 22.0 & 25.3 & \textit{30.5} & \textbf{30.9} & 29.4 & 29.3 \\
& \multirow{2}{*}{R@100} & Y & 24.5 & 27.6 & \textbf{30.3} & 29.8 & \textit{30.0} & 29.9 \\
&& N & 27.4 & 30.9 & \textit{35.8} & \textbf{35.8} & 35.1 & 35.0 \\
\hline
\multirow{8}{*}{\textsc{SGCls}} & \multirow{2}{*}{mR@50} & Y & 5.8 & 6.8 & 7.1 & 9.4 & \textit{9.6} & \textbf{12.7} \\
&& N & 12.1 & 13.5 & 15.4 & 19.8 & \textit{21.4} & \textbf{25.6} \\
& \multirow{2}{*}{mR@100} & Y & 6.0 & 7.8 & 7.6 & 10.0 & \textit{10.2} & \textbf{13.4} \\
&& N & 16.9 & 19.6 & 20.6 & 26.2 & \textit{29.1} & \textbf{32.1} \\
& \multirow{2}{*}{R@50} & Y & 34.6 & 32.4 & 35.8 & 36.7 & \textbf{38.0} & \textit{37.3} \\
&& N & 43.4 & 40.5 & 44.5 & 45.9 & \textbf{47.7} & \textit{46.9} \\
& \multirow{2}{*}{R@100} & Y & 35.4 & 34.0 & 36.5 & 37.4 & \textbf{38.8} & \textit{38.0} \\
&& N & 47.2 & 43.7 & 47.7 & 49.0 & \textbf{51.1} & \textit{50.3} \\
\hline
\multirow{8}{*}{\textsc{PredCls}} & \multirow{2}{*}{mR@50} & Y & 9.8 & 13.3 & 13.3 & 17.7 & \textit{19.3} & \textbf{22.1} \\
&& N & 20.3 & 24.8 & 27.5 & 36.3 & \textit{41.1} & \textbf{44.5} \\
& \multirow{2}{*}{mR@100} & Y & 10.5 & 15.8 & 14.4 & 19.2 & \textit{20.9} & \textbf{24.0} \\
&& N & 28.9 & 37.3 & 37.9 & 49.0 & \textit{55.4} & \textbf{58.7} \\
& \multirow{2}{*}{R@50} & Y & 59.3 & 59.9 & 65.2 & 65.8 & \textbf{66.6} & \textit{66.6} \\
&& N & 75.2 & 71.3 & 81.1 & 81.9 & \textbf{83.6} & \textit{83.5} \\
& \multirow{2}{*}{R@100} & Y & 61.3 & 64.1 & 67.1 & 67.6 & \textbf{68.2} & \textit{68.2} \\
&& N & 83.6 & 81.2 & 88.3 & 88.9 & \textbf{90.5} & \textit{90.3} \\
\hline
\end{tabular}
\end{center}
\end{table}

%% file: tables/ablation.tex
\begin{table}[t]
\begin{center}
\caption{\label{table:ablation} Ablation study on Visual Genome. All numbers are in percentage, and graph constraint is enforced.} 
\begin{tabular}{c|c c c c | c c c c}
\hline
\multirow{2}{*}{Method} & \multicolumn{4}{c|}{\textsc{SGGen}} & \multicolumn{4}{c}{\textsc{PredCls}}  \\
& mR@50 & mR@100 & R@50 & R@100 & mR@50 & mR@100 & R@50 & R@100 \\
\hline\hline
No Knowledge & 5.5 & 6.6 & 25.3 & 28.8 & 15.4 & 16.8 & 62.5 & 64.5 \\
$T=1$ & 5.6 & 6.7 & 24.9 & 28.5 & 15.6 & 17.1 & 62.1 & 64.2 \\
$T=2$ & 5.7 & 6.9 & 26.1 & 29.7 & 18.2 & 19.7 & 66.7 & 68.4 \\
\textsc{GB-Net} & \textbf{6.1} & \textbf{7.3} & \textbf{26.4} & \textbf{30.0} & \textbf{18.2} & \textbf{19.7} & \textbf{67.0} & \textbf{68.6} \\
\hline
\end{tabular}
\end{center}
\end{table}

%% file: 5_conclusion.tex
\section{Conclusion\label{sec:conclusion}}

We proposed a new method for Scene Graph Generation that incorporates external commonsense knowledge in a novel, graphical neural framework. We unified the formulation of scene graph and commonsense graph as two types of knowledge graph, which are fused into a single graph through a dynamic message passing and bridging algorithm. Our method iteratively propagates messages to update nodes, then compares nodes to update bridge edges, and repeats until the two graphs are carefully connected. Through extensive experiments, we showed our method outperforms the state of the art in various metrics.

\noindent\textbf{Acknowledgement}
This work was supported in part by Contract N6600119C4032 (NIWC and DARPA). The views expressed are those of the authors and do not reflect the official policy of the Department of Defense or the U.S. Government.

%% file: supp/1_introduction.tex
\begin{figure*}[h]
\begin{center}
\includegraphics[width=.99\linewidth]{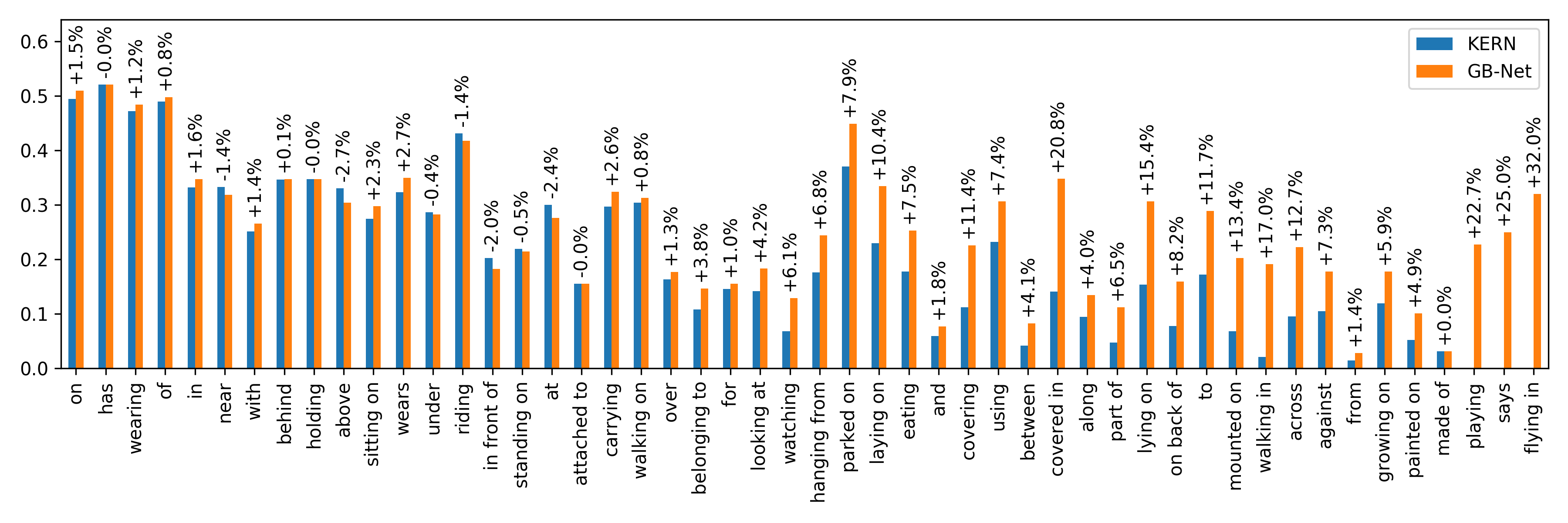}
\end{center}
   \caption{Comparison of our method \textsc{GB-Net} with KERN \cite{chen2019knowledge} in terms of recall at 50 per predicate class, without graph constraint. The horizontal axis was ordered decreasingly based on frequency in VG.}
\label{fig:perclass}
\end{figure*}

In this document, we provide additional details that were omitted from the main manuscript due to the space constraint. We start by an analysis of our quantitative results, were we show our method addresses the shortcoming of the state of the art in modeling the tail of the predicate distribution. We then report an empirical analysis of the computational cost of our method, showing it is significantly faster than the state of the art, while also more accurate. After that, we clearly describe the process of creating our commonsense graph, and we provide a set of qualitative examples to illustrate how our method exploits such commonsense knowledge. We conclude by describing the code that we will provide to reproduce the results. 

%% file: supp/2_visualization.tex
\section{Per-class performance\label{sec:viz}}

Figure~\ref{fig:perclass} illustrates the recall of our method for each predicate class separately, where predicates are ordered decreasingly according to their frequency in the training set. 
While state-of-the-art methods such as KERN~\cite{chen2019knowledge} obtain much lower performance on the tail of the distribution, our method significantly improves the performance of the tail without losing on the frequent predicates, resulting in a more reliable 
and consistent 
performance overall. 

%% file: supp/3_time.tex
\section{Computational cost\label{sec:time}}

We compute the training and test speed of our method and compare to KERN \cite{chen2019knowledge} using identical hardware, with one GPU of type NVIDIA GeForce GTX 1080 Ti with 11 gigabytes of memory, and summarize the results in Table~\ref{table:time}.
Perhaps the most important aspect of computation is the run time when deploying the model on new images. 
To this end, we run each trained model on the entire test set of Visual Genome (VG), \textit{i.e.} 26446 images, and get the average run time over all images in terms of seconds. 
Our method is 34\% faster than the state of the art, while being significantly more accurate as demonstrated in Table 1 of the main paper.

Another important factor is the duration of training. We record the time it takes to train each model on one epoch of the VG training set, \textit{i.e.} 56224 images, and get the average over 10 training epochs. 
As Table~\ref{table:time} shows, our method is more than twice faster than the state of the art. 
One of the reasons is that KERN has two stages of message passing, each with three steps, first to infer entities, and then to infer predicates, while our method infers both entities and predicates jointly, through 3 steps of global message passing. 

Finally, we compare the number of trainable parameters each method has. 
Our method has 10\% more parameters than KERN, while it is 52\% and 34\% faster than KERN during training and test respectively. 
Note that in all methods, 139.8 millions of the parameters belong to the Faster R-CNN detector, which we fix while training for scene graph generation.

\begin{table}[t]
\begin{center}
\caption{\label{table:time}Time and memory cost of our method compared to the state of the art}
\begin{tabular}{|c|c|c|c|}
\hline
\multirow{2}{*}{Method} & Test time & Train time & \# parameters \\
& (sec/image) & (min/epoch) & (million) \\
\hline\hline
KERN \cite{chen2019knowledge} & 0.79 & 401.2 & 405.2 \\
\textsc{GB-Net} & 0.52 & 191.6 & 444.6 \\
\hline
\end{tabular}
\end{center}
\end{table}

%% file: supp/4_graph.tex
\section{Commonsense graph construction\label{sec:graph}}

Our method utilizes a background knowledge graph which encodes commonsense knowledge about the target entity and predicate classes. 
Such information have been proved to be effective for scene graph generation~\cite{chen2019knowledge,gu2019scene}. 
Intuitively, this is because they can help the model disambiguate between possible visual classes, using higher-level semantic meanings and relationships of classes. 
For instance, object affordance is a type of commonsense, because a typical human knows a \texttt{Bike} can be used for \texttt{riding}. 
This fact can be used by the model when it perceives a bounding box of \texttt{Person} above a bounding box of \texttt{Bike}, and it is not sure whether to classify their relationship as \texttt{riding} or \texttt{onTopOf} or \texttt{mountedOn}.

Commonsense does not have to be homogeneous. Besides affordance, ontological hierarchy is another type of useful information for scene graph generation. A typical human knows \texttt{Man} and \texttt{Woman} are both subtypes of \texttt{Person}, and hence can generalize the properties of \texttt{Man} to \texttt{Woman}. 
Another example is statistical commonsense, where an average human knows a \texttt{Vase} is more likely to be \texttt{on} a \texttt{Table} than anywhere else, not because of its semantic characteristics, but because it is usually that way. 

Our method is independent of the content of the commonsense graph, as long as its nodes are exactly the set of classes required by the target task, and every fact is encoded as a directed, typed, weighted edge. 
Accordingly, our commonsense graph has 150 entities and 50 predicates as conventionally used in the SGG task, and we compile the edges (\textit{i.e.} facts) from a variety of sources, to make it as rich and comprehensive as possible. 
To this end, we first manually find for each of the 200 VG classes, the WordNet Synset~\cite{miller1995wordnet} that represents its meaning. 
This step is required to fix a deterministic meaning for each class, such that for instance, \texttt{Cabinet} always represents the furniture piece and is not confused with its meaning in politics.

Given the 200 nodes that are grounded on WordNet, we start collecting commonsense facts to encode as edges. 
The most prominent repository of commonsense knowledge is ConceptNet~\cite{liu2004conceptnet}, which is a large-scale graph where nodes are \emph{concepts}, including but not limited to entity and predicate classes, and edges encode relational facts about pairs of concepts. There are over 21 million edges of 34 types connecting over 8 million nodes in ConceptNet. We query each of our 200 nodes to find its semantically closest node in ConceptNet, by using the WordNet Synset title and the ConceptNet search API. Then we query all ConceptNet edges between each pair of those 200 nodes, 
and further manually clean and prune those edges, leading to 104 edges of 5 types. Those 5 edge types are: \texttt{partOf} as in \texttt{Hand-partOf-Person}, \texttt{relatedTo} as in \texttt{Lamp-relatedTo-Light}, \texttt{isA} as in \texttt{Dog-isA-Animal}, \texttt{mannerOf} as in \texttt{MountedOn-mannerOf-AttachedTo}, and \texttt{usedFor} as in \texttt{Bike-usedFor-Riding}. Further, for asymmetric relationships (all except \texttt{relatedTo}), we create a reverse edge with a distinct type. For instance, because we have \texttt{Hand}-\texttt{partOf}-\texttt{Person}, we also create the fact \texttt{Person}-\texttt{hasPart}-\texttt{Hand}. Our message passing framework, unlike conventional ones, only propagates messages in the direction of each edge, and not backwards. Hence, the way the \texttt{Person} node is affected by \texttt{Hand} would be different from how \texttt{Hand} is affected by \texttt{Person}, because we do not share parameters across edge types. Since there is no edge confidence in ConceptNet, these edges all have a binary weight ($1.0$ if exists and $0.0$ if not).

Besides ConceptNet, we also use WordNet to get ontological similarity of words. Although WordNet is not originally identified as a commonsense graph, the knowledge we extract from it is trivial, generally known information, and hence can be considered commonsense. 
Inspired in part by~\cite{kato2018compositional}, we use three similarity metrics of the WordNet API (namely path similarity, Leacock Chordorow (LCH) similarity, and Wu-Palmer (WUP) similarity \cite{miller1995wordnet}), and a manually tuned threshold for each, to determine whether two entity classes are relevant or not. This is encoded in the edge type \texttt{WordNetSimilarTo} with binary weights. This strategy does not work well for predicate classes, so this edge is only between pairs of entities. 

Finally, we use the VG training set to get co-occurrence statistics between categories, inspired by~\cite{zellers2018neural} but in a more comprehensive manner. 
We estimate conditional probabilities of subject given predicate, object given predicate, predicate given subject, predicate given object, subject given object, and object given subject, as well as the correlation of entity classes as they connect to the same predicate, and the correlation of predicate classes as they connect to the same entity. These edge types capture a variety of statistical interactions between classes. 
Each of those statistics result in a pairwise matrix, which we then sparsify by keeping the top 5 element in each row and setting the rest to zero. For instance, \texttt{riding} is connected to its top 5 most likely objects, namely \texttt{Horse}, \texttt{Bike}, \texttt{Skateboard}, \texttt{Motorcycle} and \texttt{Wave}. Note that although \texttt{riding} is not connected to \texttt{Elephant}, its 6th most likely object, \texttt{Elephant} and \texttt{Horse} are both connected to \texttt{Animal} through \texttt{isA} edges, and they are connected to each other through \texttt{WordNetSimilarTo}, allowing our message passing framework to exploit this rich structure and infer \texttt{Elephant} can be ridden too.

Overall, these three sources lead to 19 edge types (including backward edge types for asymmetric relationships). Carefully compiled from multiple sources, our commonsense graph is more sophisticated and complete, compared to those made for recent knowledge-aware computer vision systems such as~\cite{chen2018iterative} and~\cite{kato2018compositional}. Note that none of those papers publicized their knowledge graph, so we were unable to compare.
The process of graph generation involves manual effort, thus we have made our commonsense graph publicly available as a part of our code.

%% file: supp/6_qualitative.tex
\section{Qualitative results\label{sec:qual}}
 
To demonstrate the performance of our method qualitatively, Figures 2-10 show examples of scene graphs generated by our method, compared to the ones generated by KERN. These examples illustrate how our method predicts more commonsensical graphs despite visual ambiguities in the scene. We observed several patterns in which our method outperforms KERN. Since KERN (and most other SGG methods such as \cite{zellers2018neural}) first classify each entity and then classify predicates, they are unable to utilize predicate semantics to enhance entity classification. Thus in many cases, KERN misclassifies an entity, due to visual ambiguity and clutter, while our method makes the correct prediction that might be less apparent visually, but lead to a more consistent scene graph semantically. In some other cases, KERN misclassifies an entity not because it is visually cluttered, but because the bounding box is too loose and covers a big portion of background. Our method is more robust to such bounding box inaccuracies, resulting a higher overall performance. Finally, in some cases entities are classified correctly by KERN, but the choice of predicates is inappropriate. Our method usually picks the correct predicate, in accordance to commonsense knowledge, such as object affordances. 
More detailed discussion can be found in each figure's caption.

\begin{figure*}
\begin{center}
\includegraphics[width=.45\linewidth]{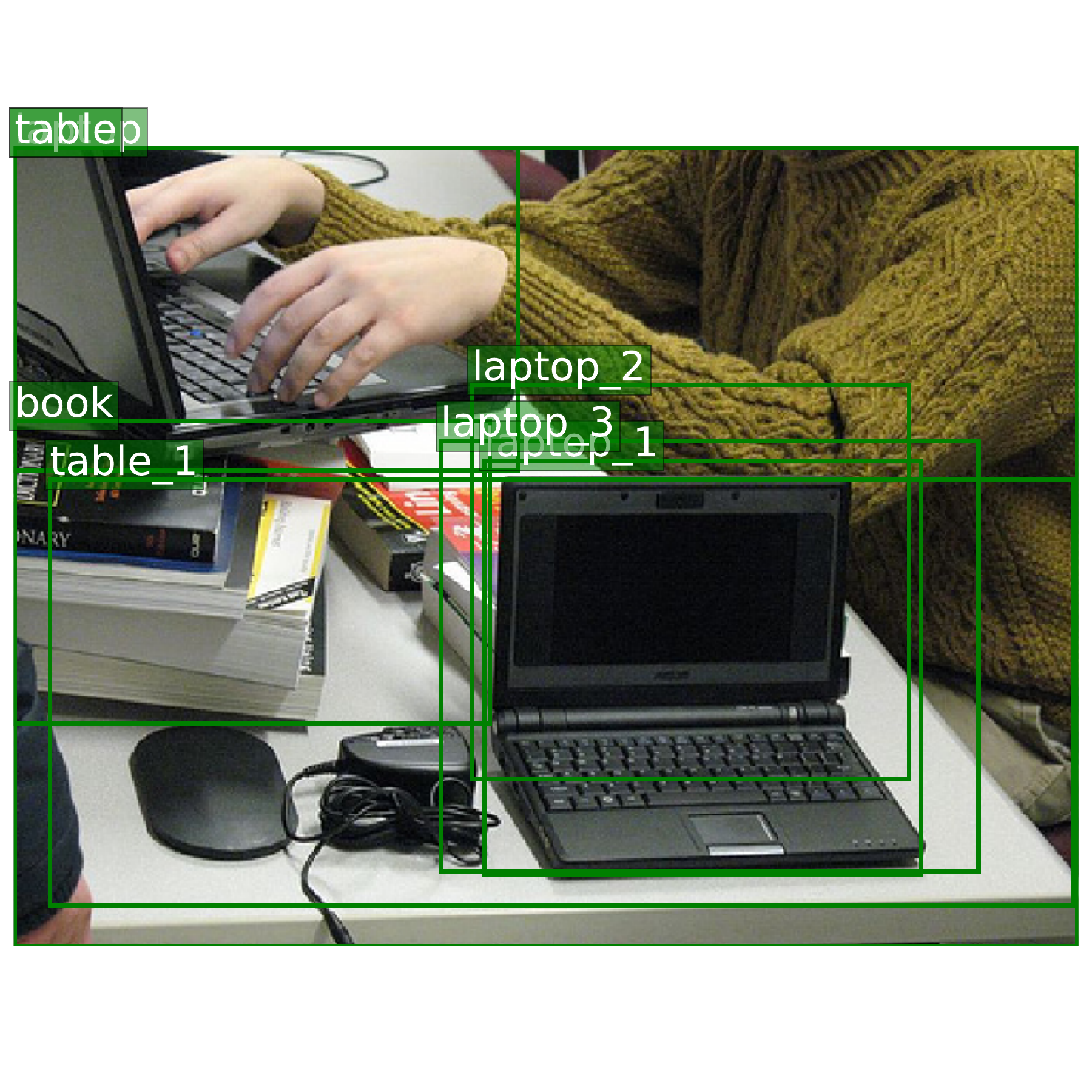}\hfill
\includegraphics[width=.45\linewidth]{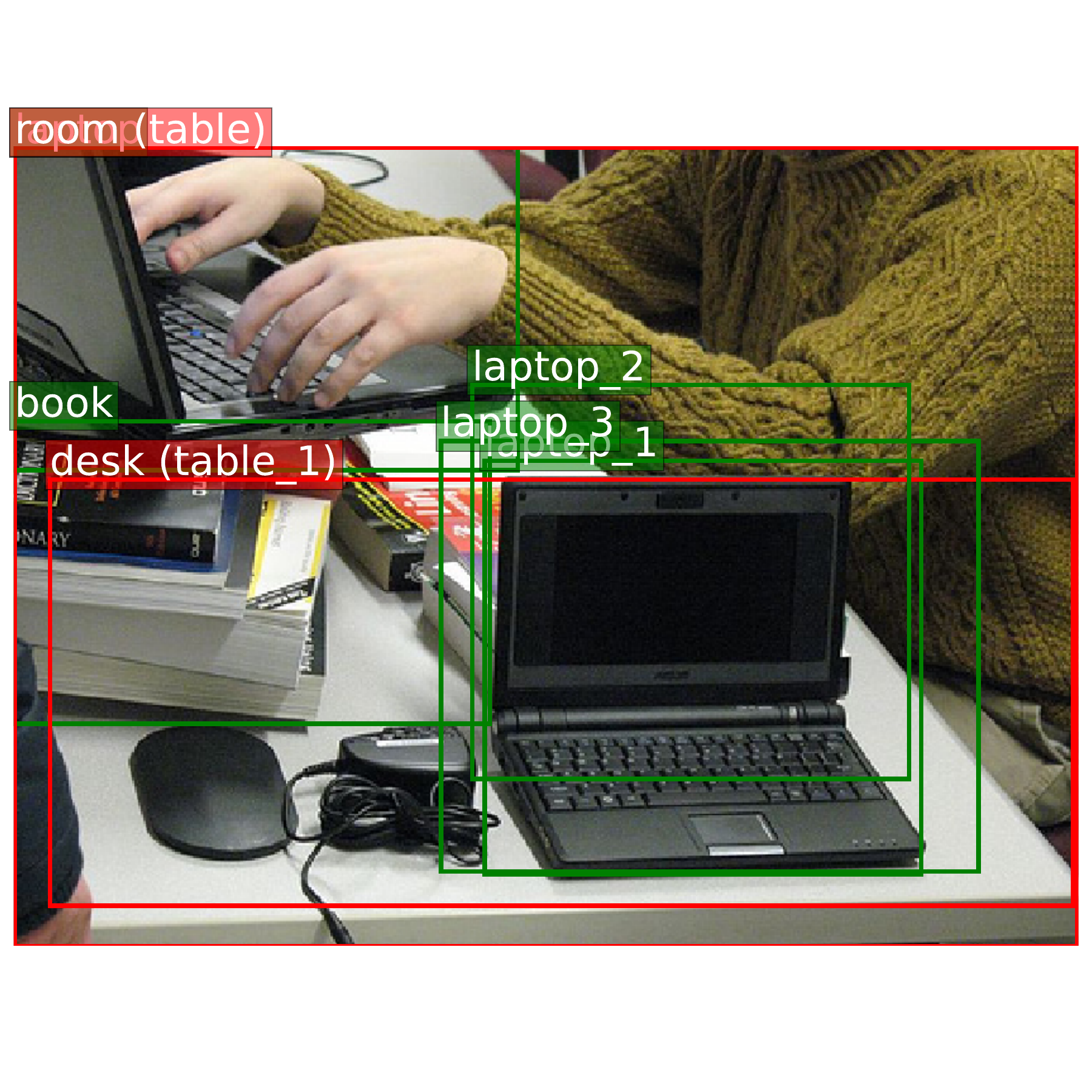}\\
\vspace{-1.0cm}
\includegraphics[width=.465\linewidth]{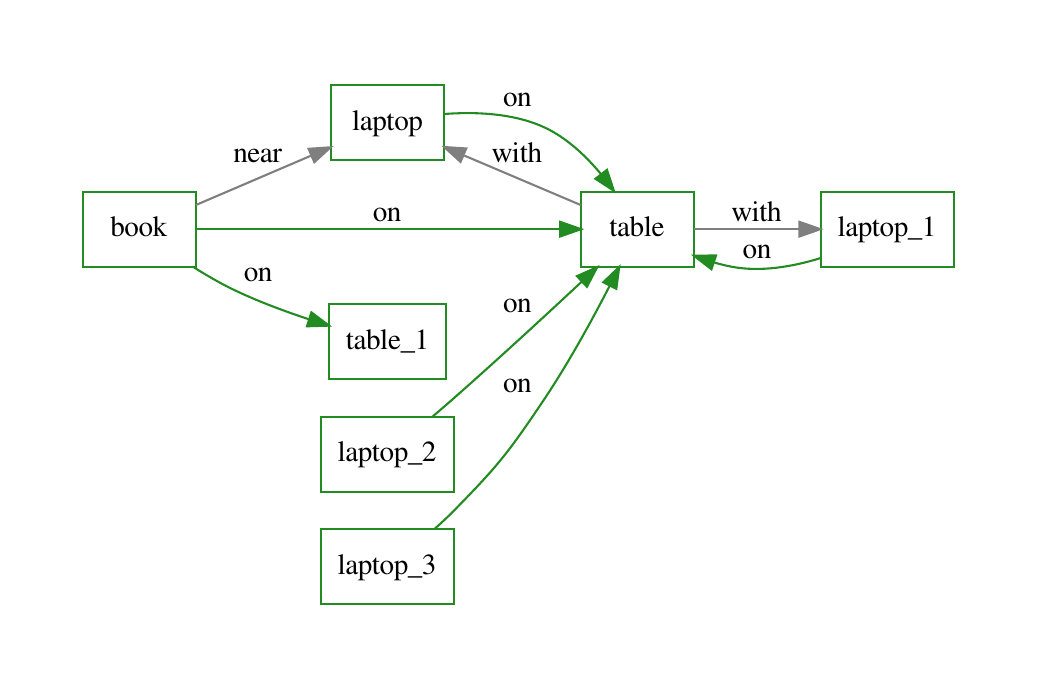}\hfill
\includegraphics[width=.535\linewidth]{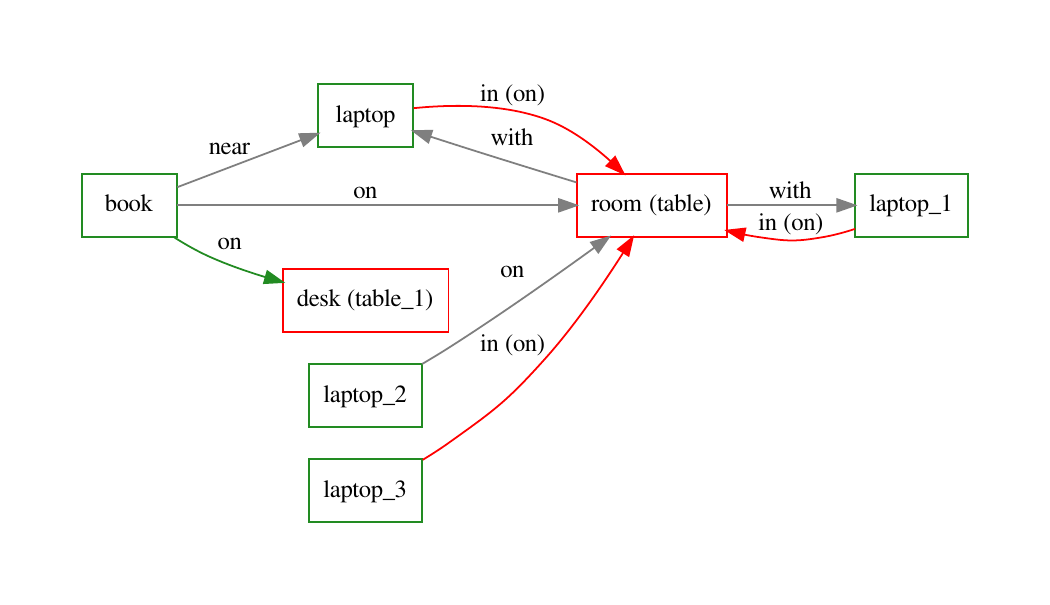}\hspace{-0.5cm}
\vspace{-0.7cm}
\end{center}
   \caption{Example comparison of our method \textsc{GB-Net} (left) with KERN \cite{chen2019knowledge} (right). Misclassified entities and predicates are colored red, and the correct class is included in parentheses. KERN misclassifies the table as a room, possibly because the bounding box contains the entire scene, but this leads to incorrect triplets such as laptop on room. Our method predicts the more appropriate class table, that makes every triplet more commonsensical.}
\label{fig:example_laptop_room}
\end{figure*}

\begin{figure*}
\begin{center}
\scalebox{.9}{
\includegraphics[width=.45\linewidth]{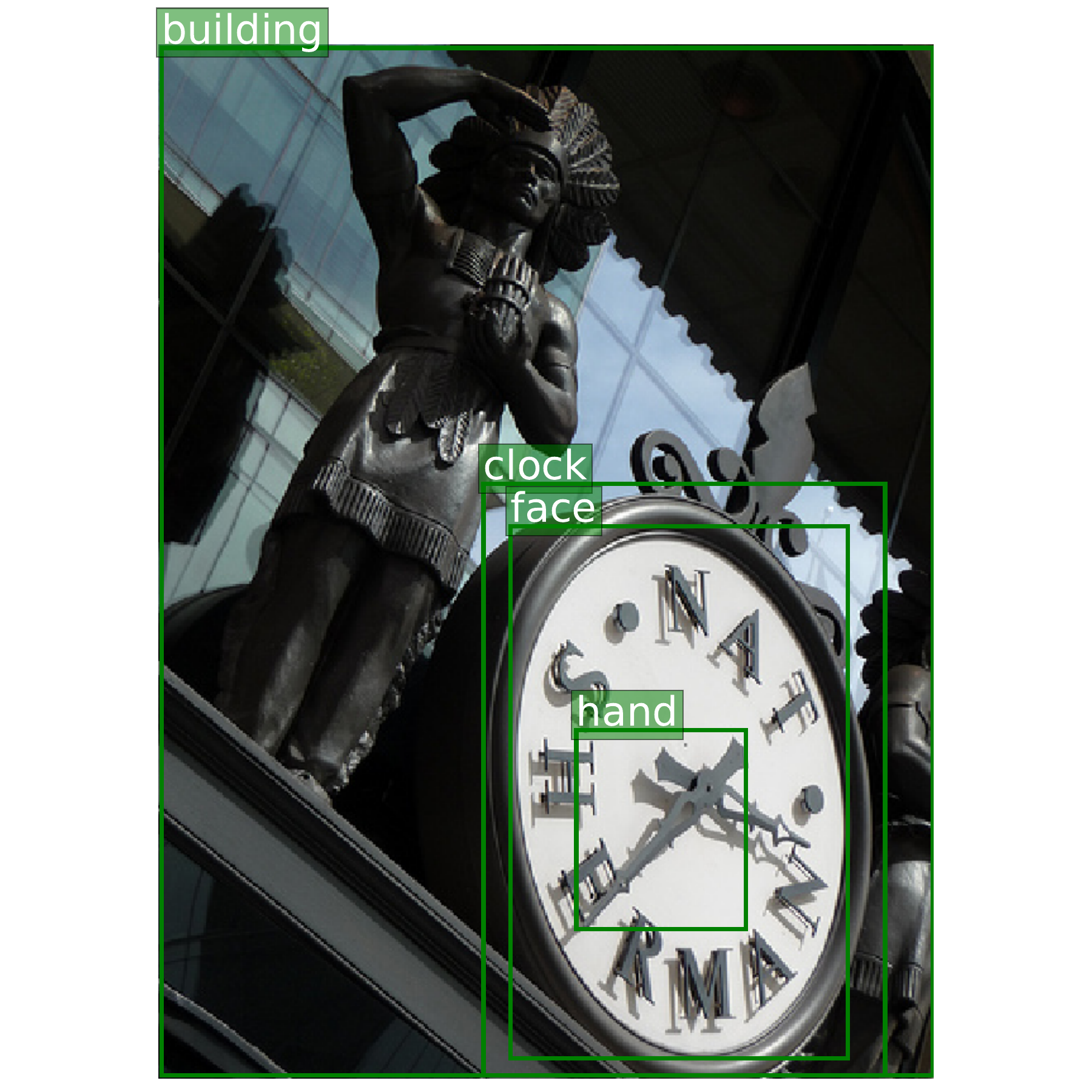}\hfill
\includegraphics[width=.45\linewidth]{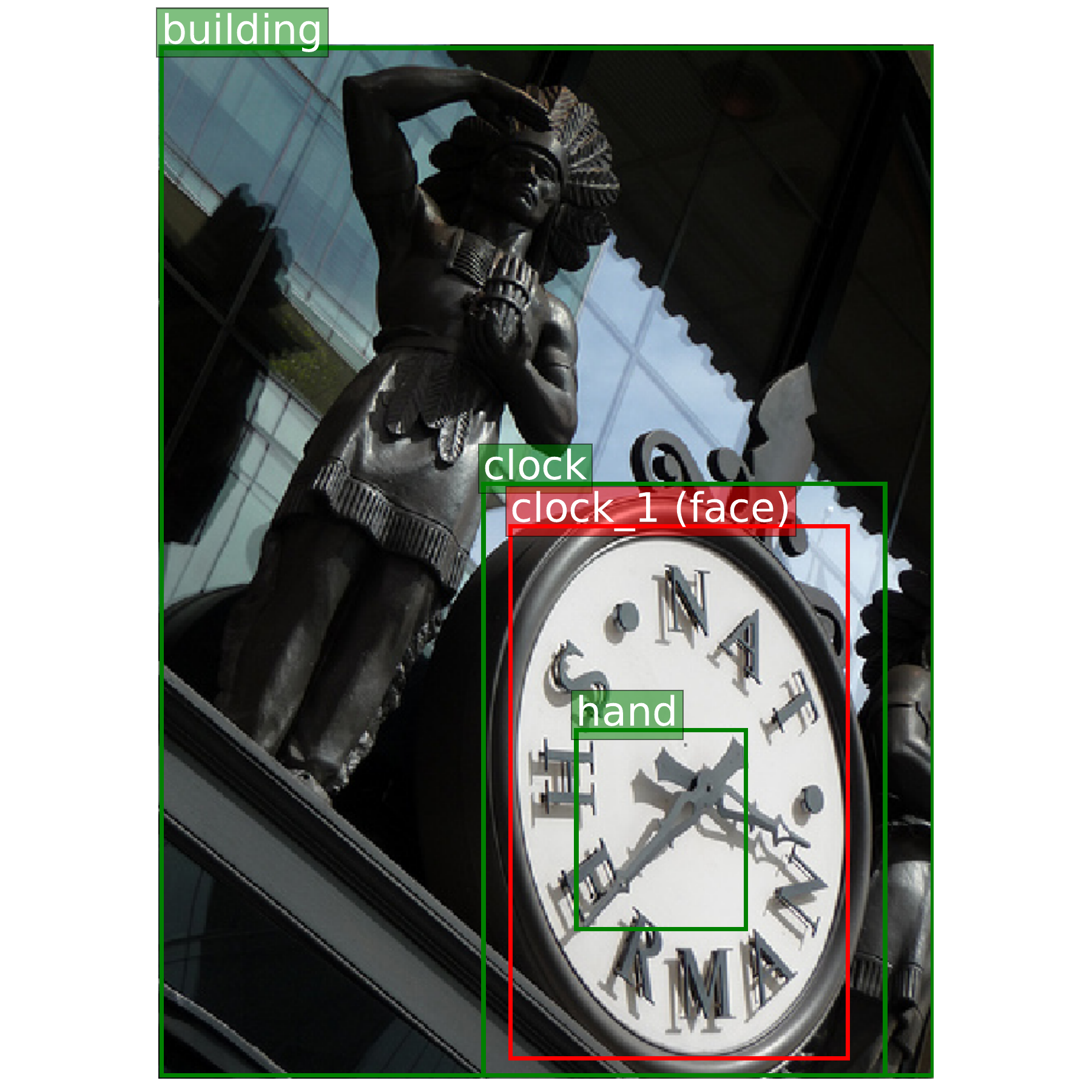}
}\\
\vspace{-0.5cm}
\includegraphics[width=.46\linewidth]{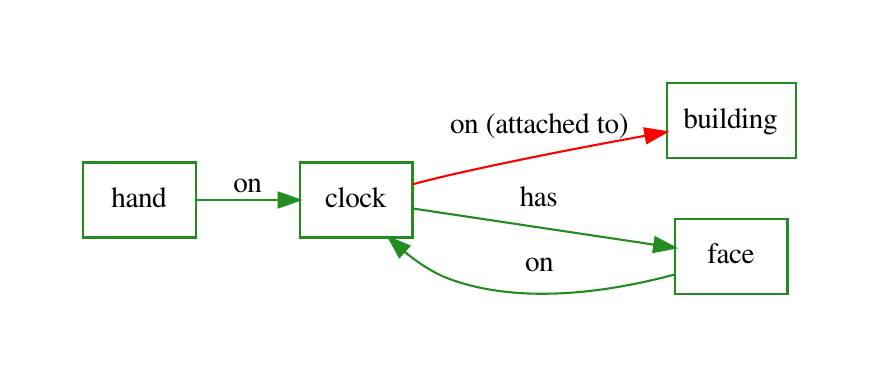}\hfill
\includegraphics[width=.5\linewidth]{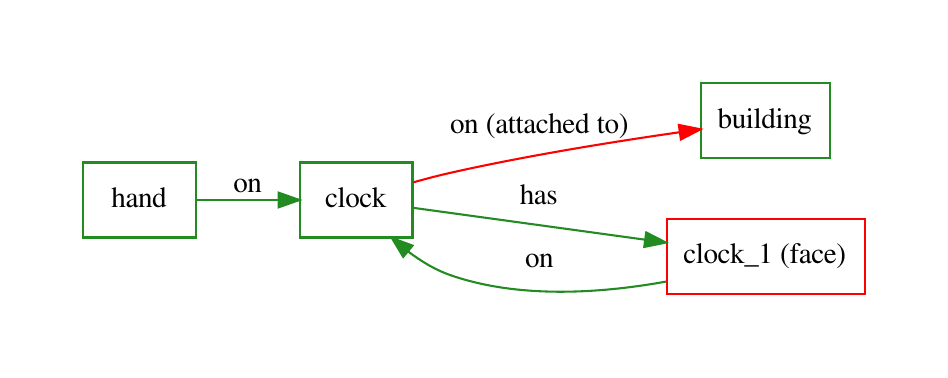}

\vspace{-0.7cm}
\end{center}
   \caption{Example comparison of our method \textsc{GB-Net} (left) with KERN \cite{chen2019knowledge} (right). The concept of a clock face is challenging for KERN but our method can produce such output, by exploiting the prior knowledge and statistics that clocks can have faces and the face would be on the clock. KERN predicts the triplet clock has clock, which does not make sense.}
\label{fig:example_clock_face}
\end{figure*}

\begin{figure*}
\begin{center}
\includegraphics[width=.45\linewidth]{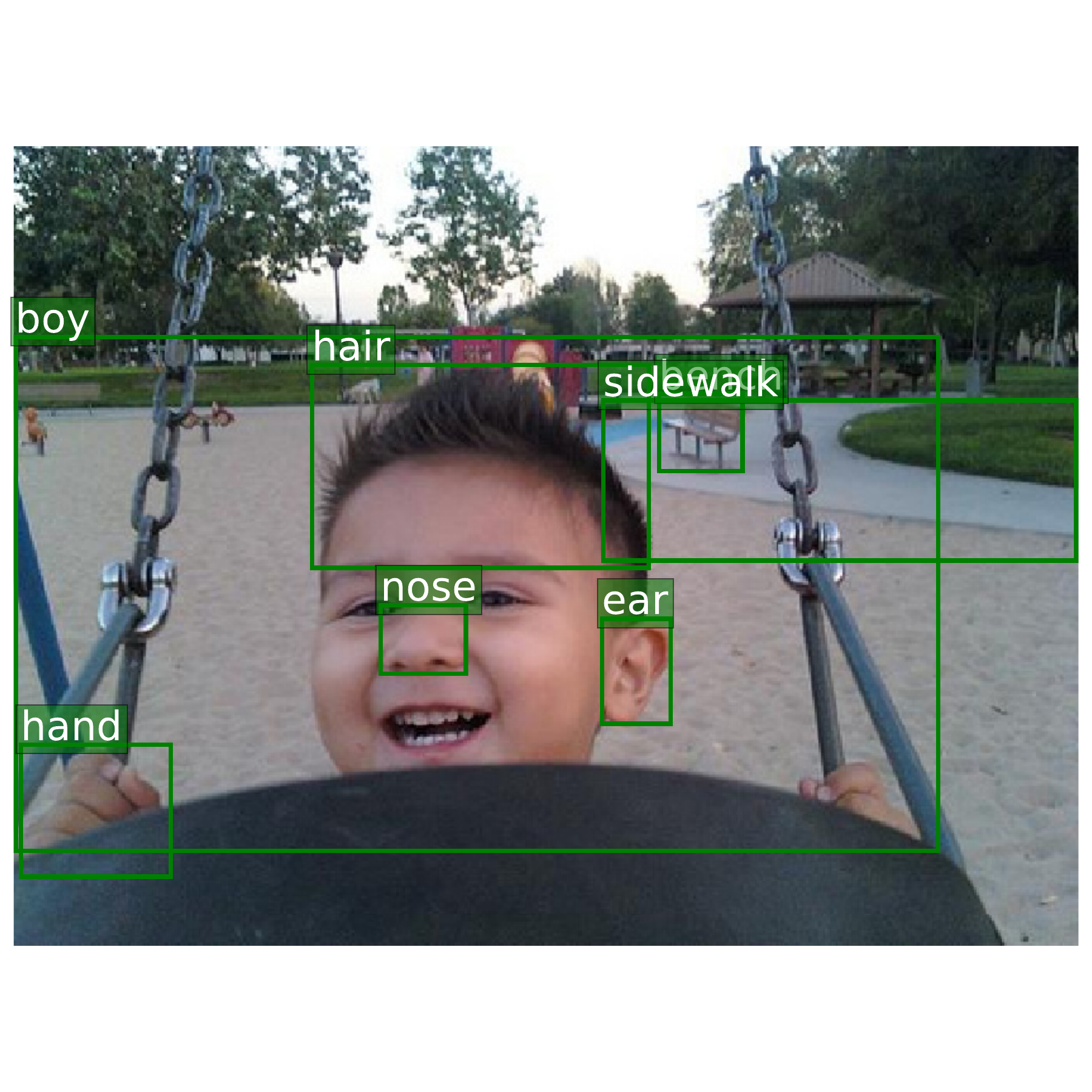}\hfill
\includegraphics[width=.45\linewidth]{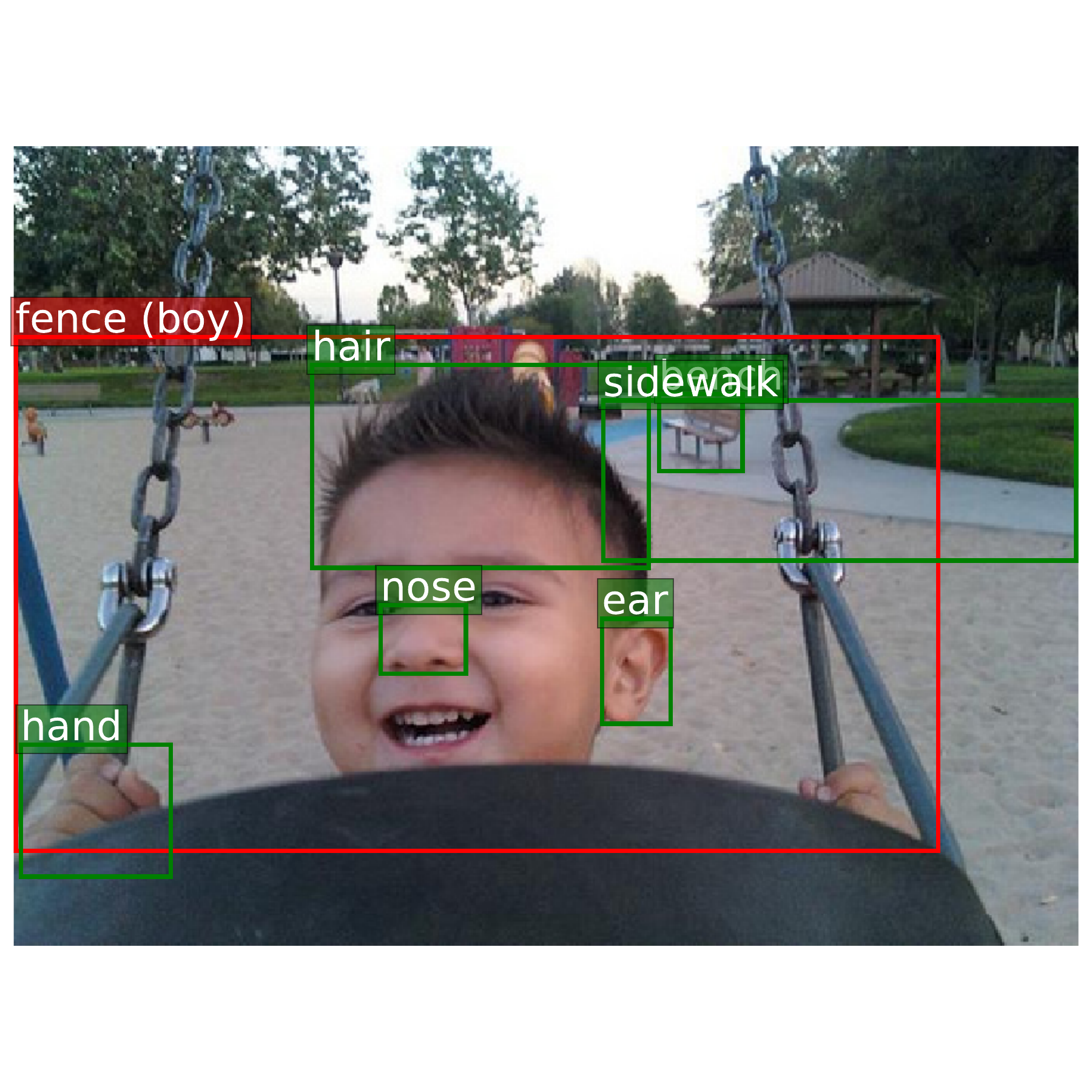}\\
\vspace{-1.0cm}
\includegraphics[width=.37\linewidth]{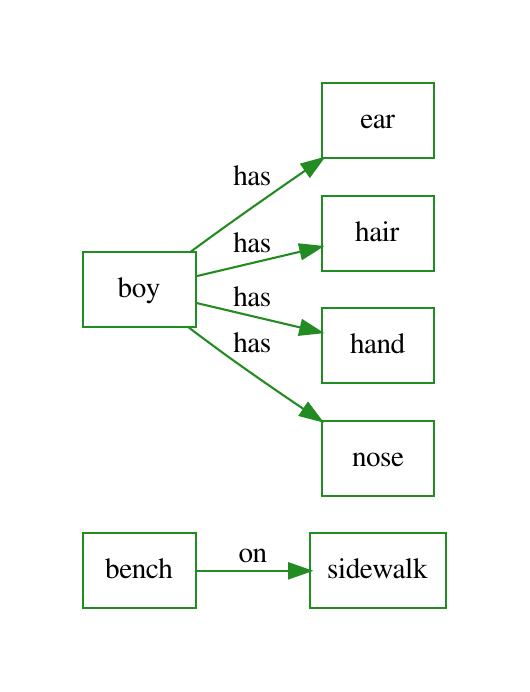}\hfill
\includegraphics[width=.48\linewidth]{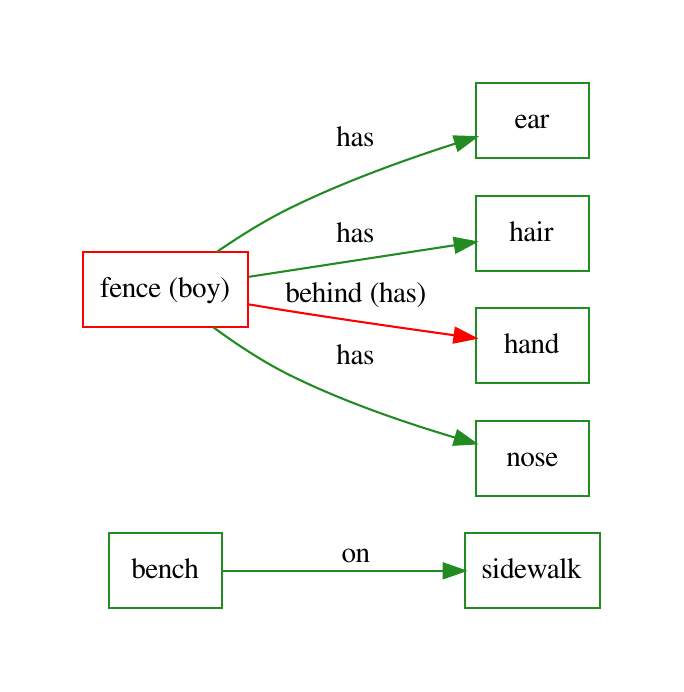}
\end{center}
   \caption{Example comparison of our method \textsc{GB-Net} (left) with KERN \cite{chen2019knowledge} (right). KERN misclassifies the boy as fence, which leads to the nonsensical triplets fence has ear, fence has nose, etc. Our method is less likely to make such meaningless predictions.}
\label{fig:example_boy_fence}
\end{figure*}

\begin{figure*}
\begin{center}
\includegraphics[width=.45\linewidth]{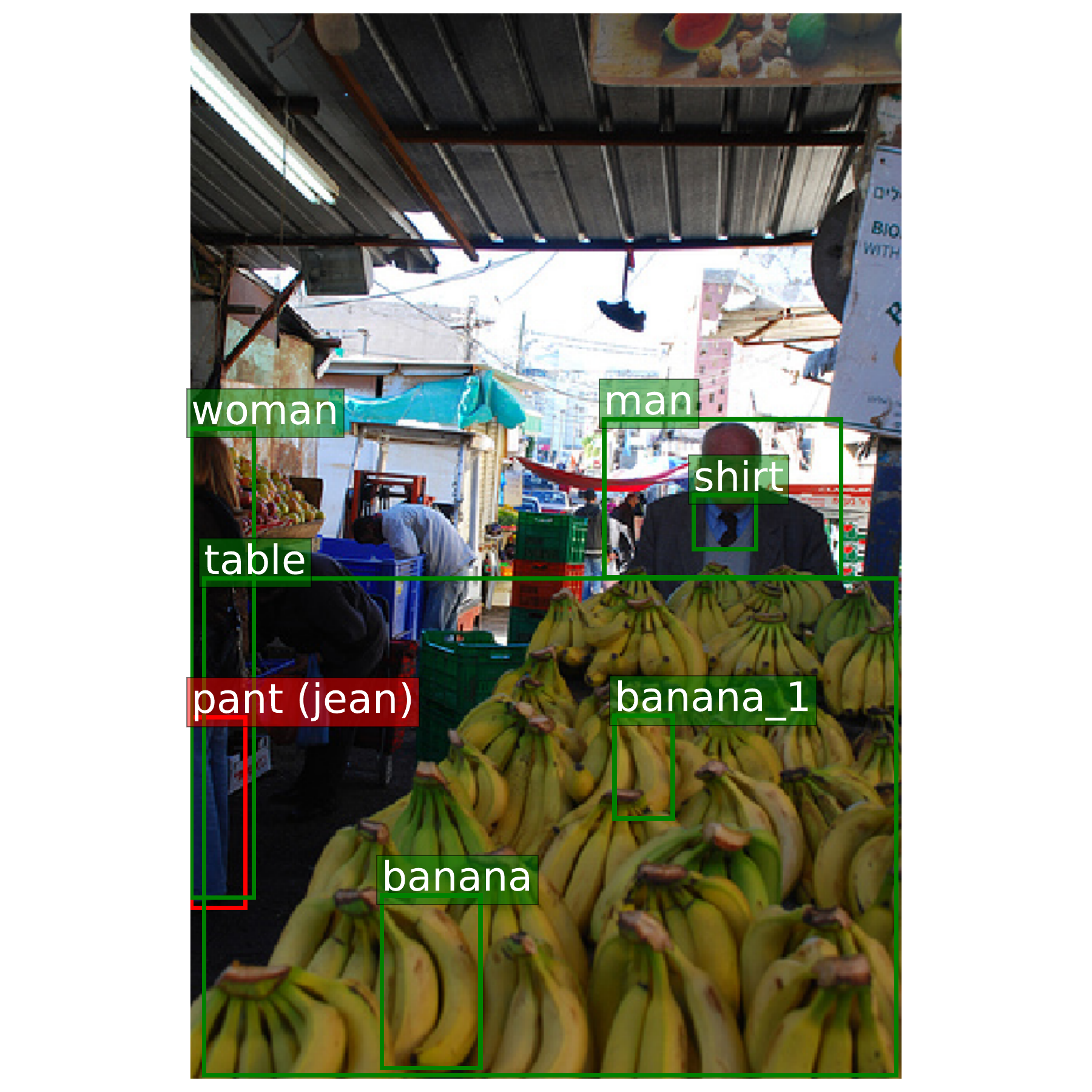}\hfill
\includegraphics[width=.45\linewidth]{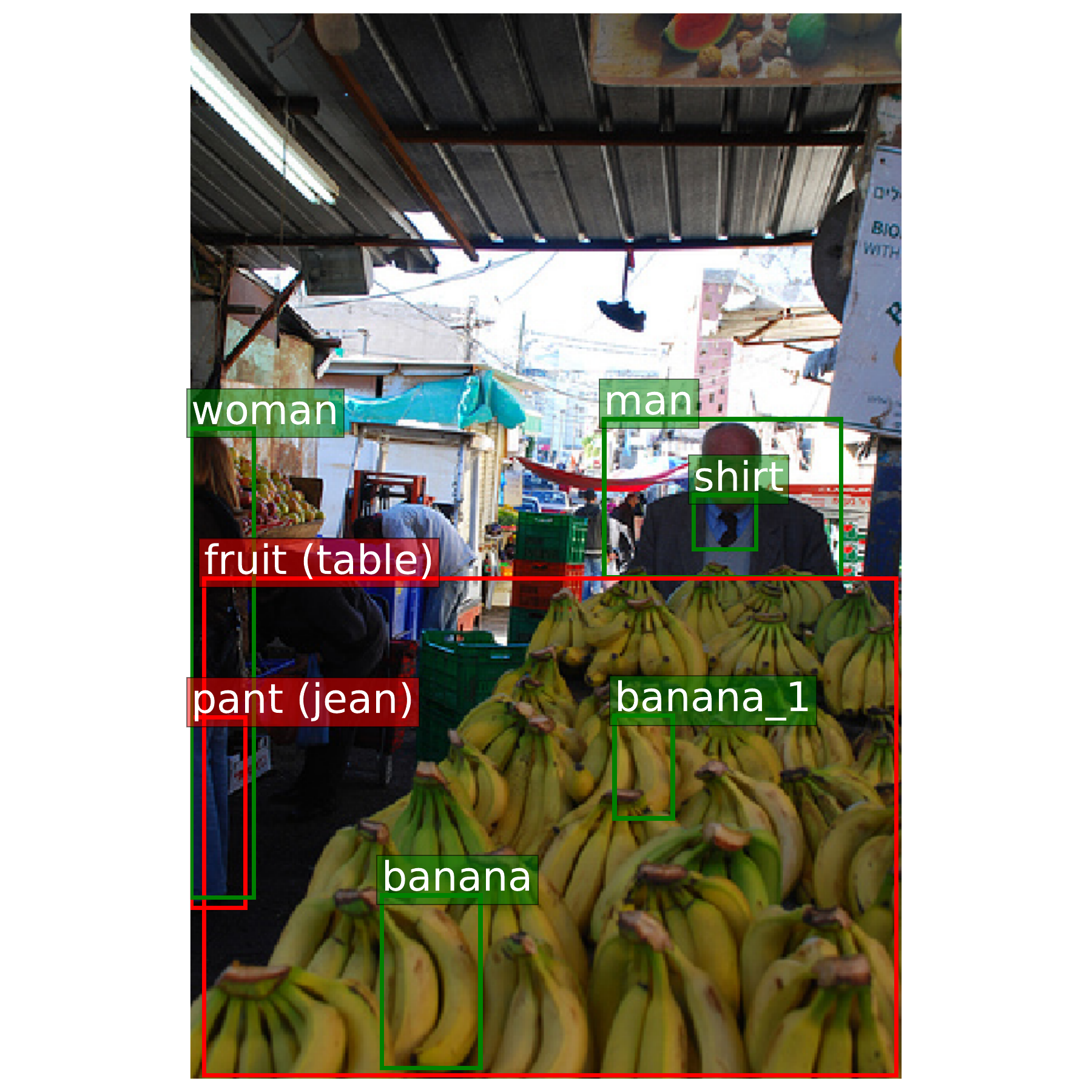}\\
\vspace{-0.8cm}
\includegraphics[width=.48\linewidth]{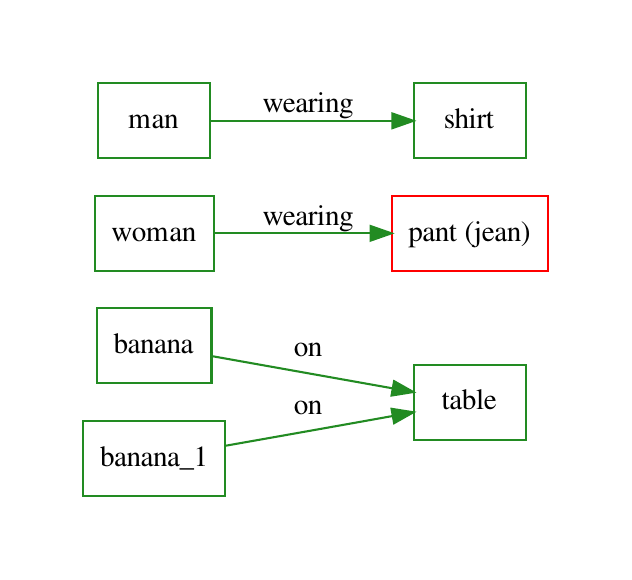}\hfill
\includegraphics[width=.49\linewidth]{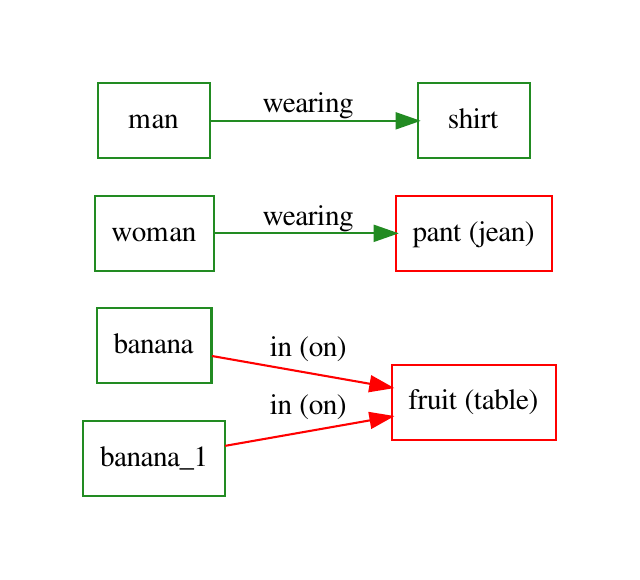}\hspace{-0.4cm}
\vspace{-1cm}
\end{center}
   \caption{Example comparison of our method \textsc{GB-Net} (left) with KERN \cite{chen2019knowledge} (right). KERN misclassifies the table as fruit, possibly because it is entirely covered by fruites. But this leads to nonsensical triplet banana in fruit. Our method correctly classifies the table, which leads to a more commonsensical scene graph.}
\label{fig:example_banana}
\end{figure*}

\begin{figure*}
\begin{center}
\scalebox{.9}{
\includegraphics[width=.45\linewidth]{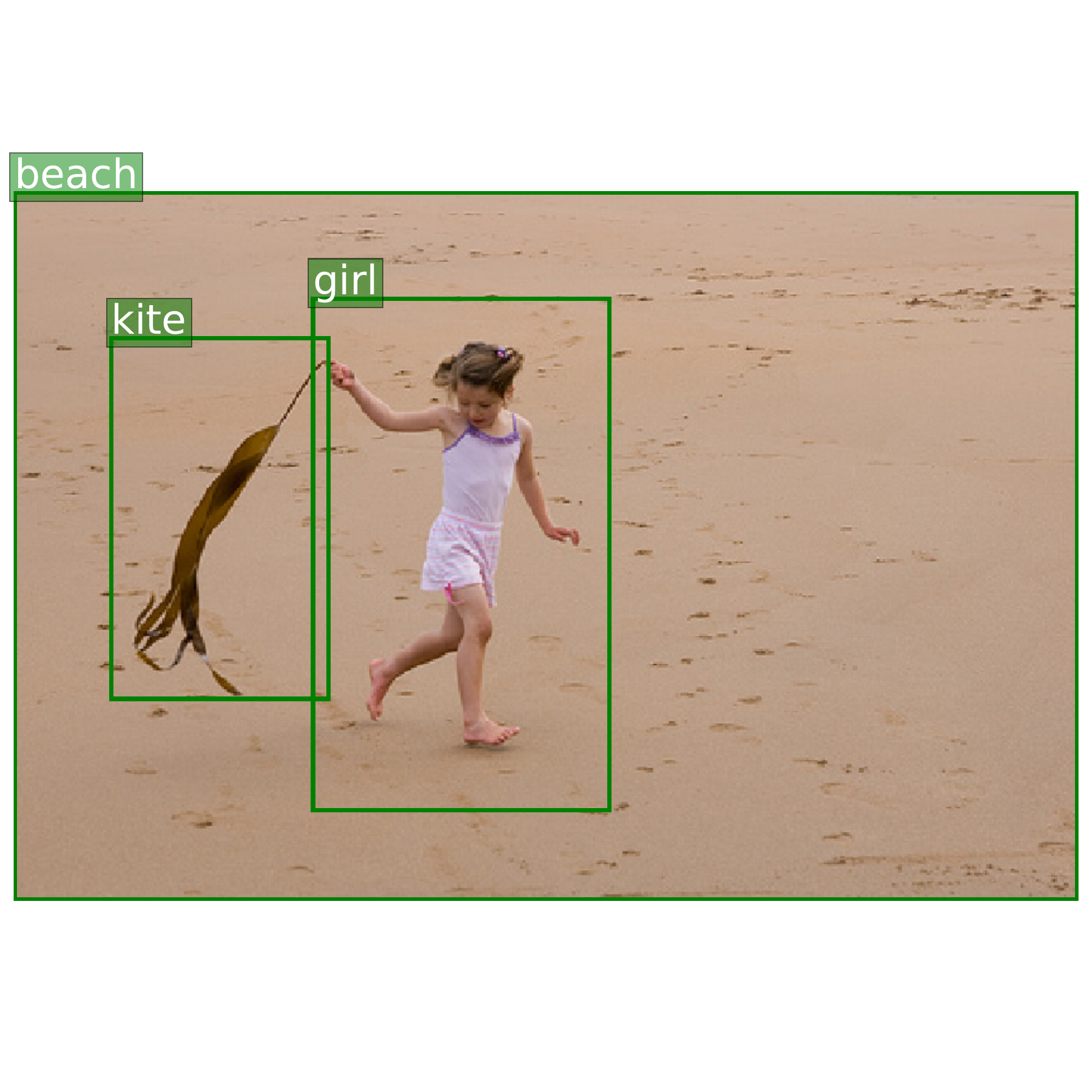}\hfill
\includegraphics[width=.45\linewidth]{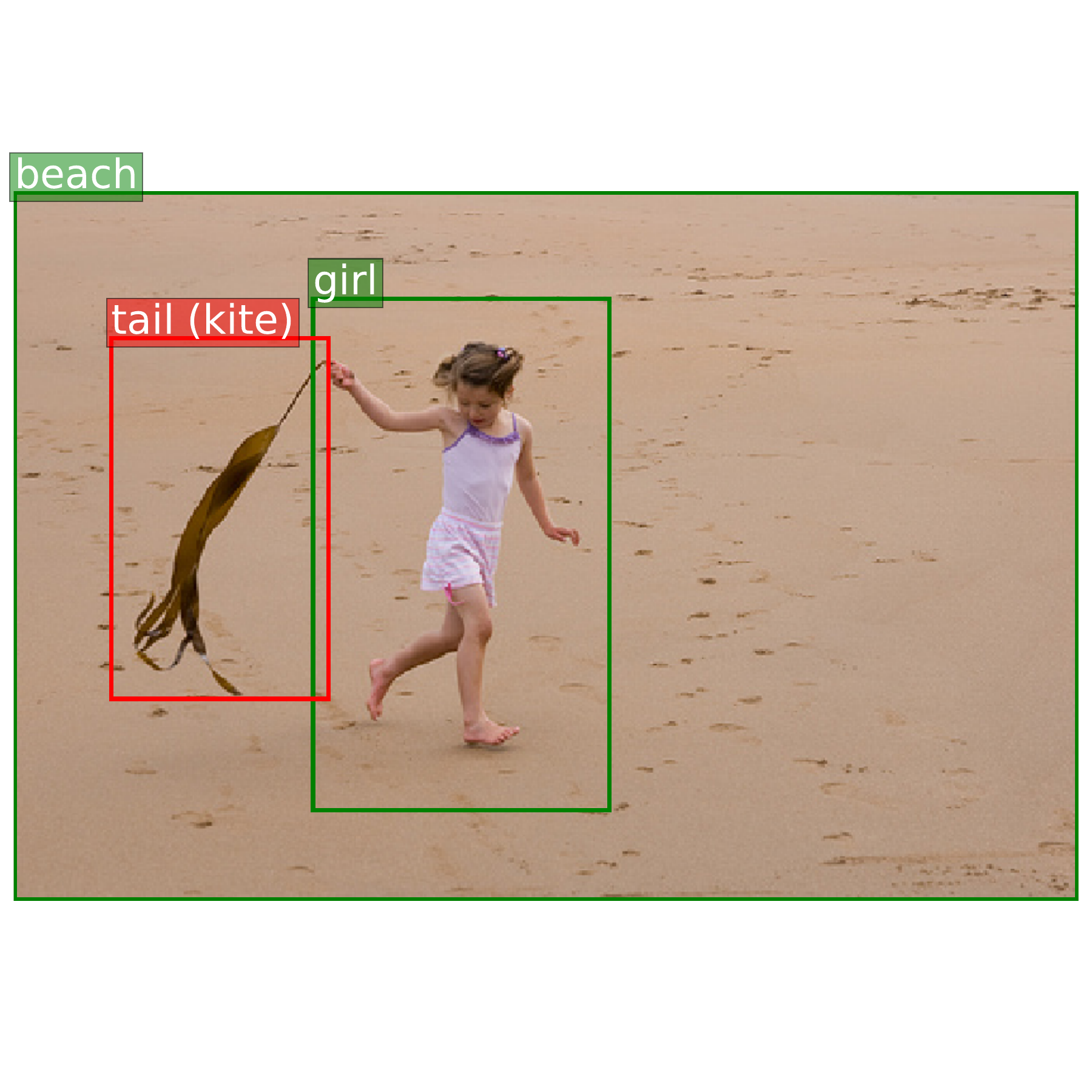}
}\\
\vspace{-1cm}
\scalebox{.8}{
\includegraphics[width=.48\linewidth]{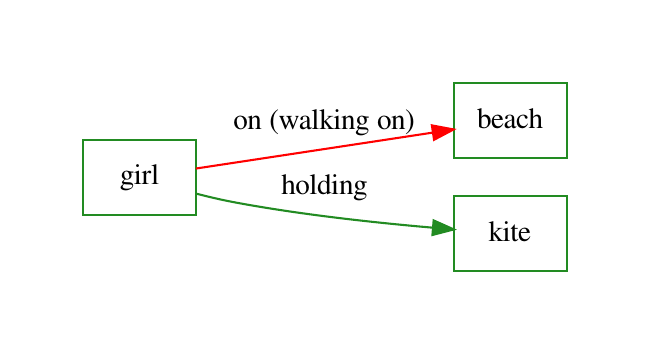}\hfill
\includegraphics[width=.48\linewidth]{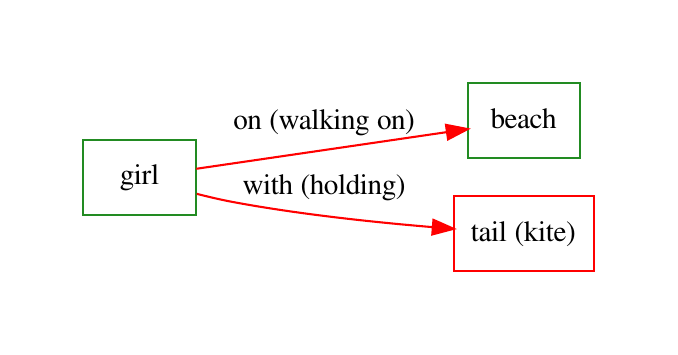}
}
\vspace{-1cm}
\end{center}
   \caption{Example comparison of our method \textsc{GB-Net} (left) with KERN \cite{chen2019knowledge} (right). KERN misclassifies the kite as a tail, because it actually looks more like a tail. Our method predicts kite that is visually less clear, but leads to a more commonsensical graph overall.}
\label{fig:example_tail}
\end{figure*}

\begin{figure*}
\begin{center}
\includegraphics[width=.45\linewidth]{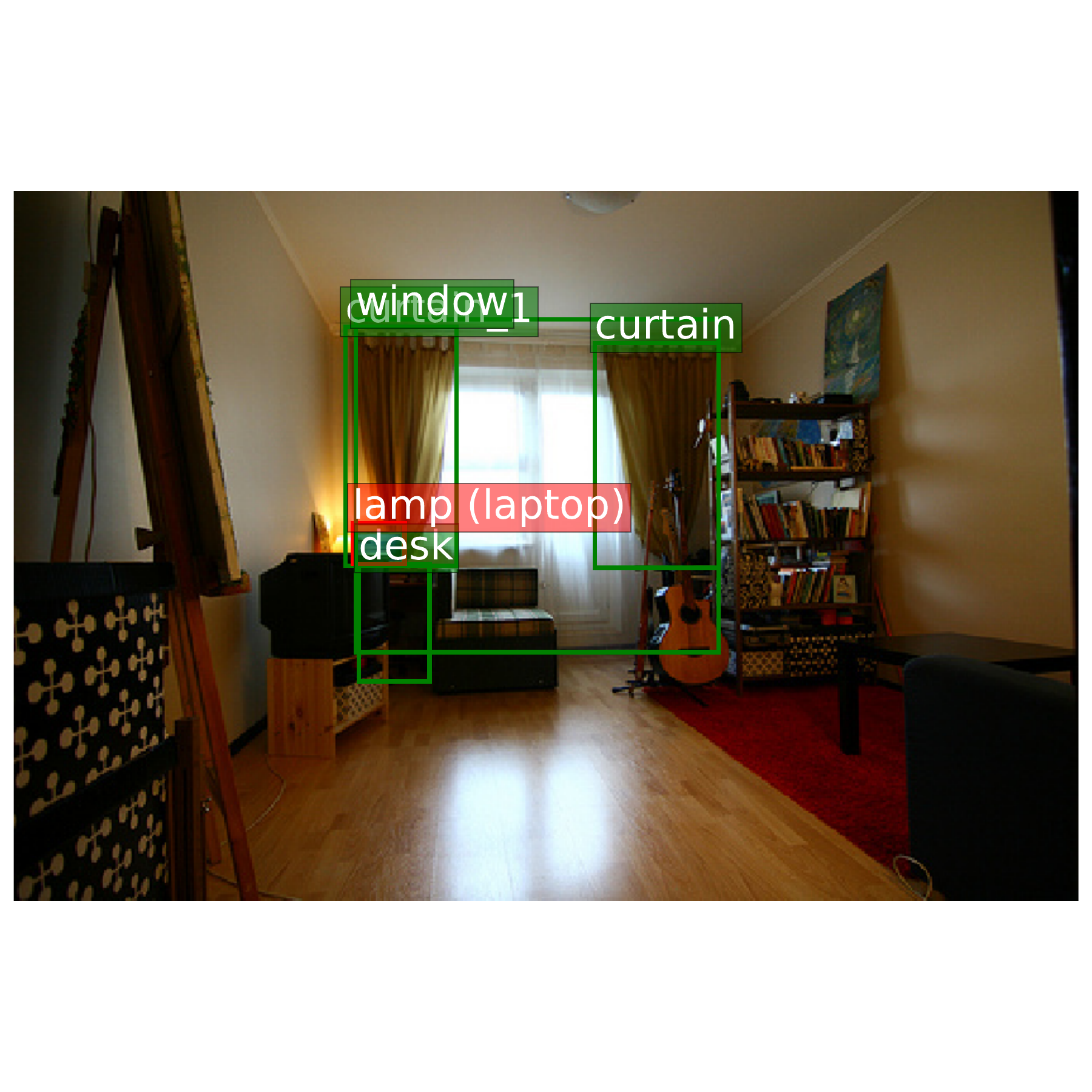}\hfill
\includegraphics[width=.45\linewidth]{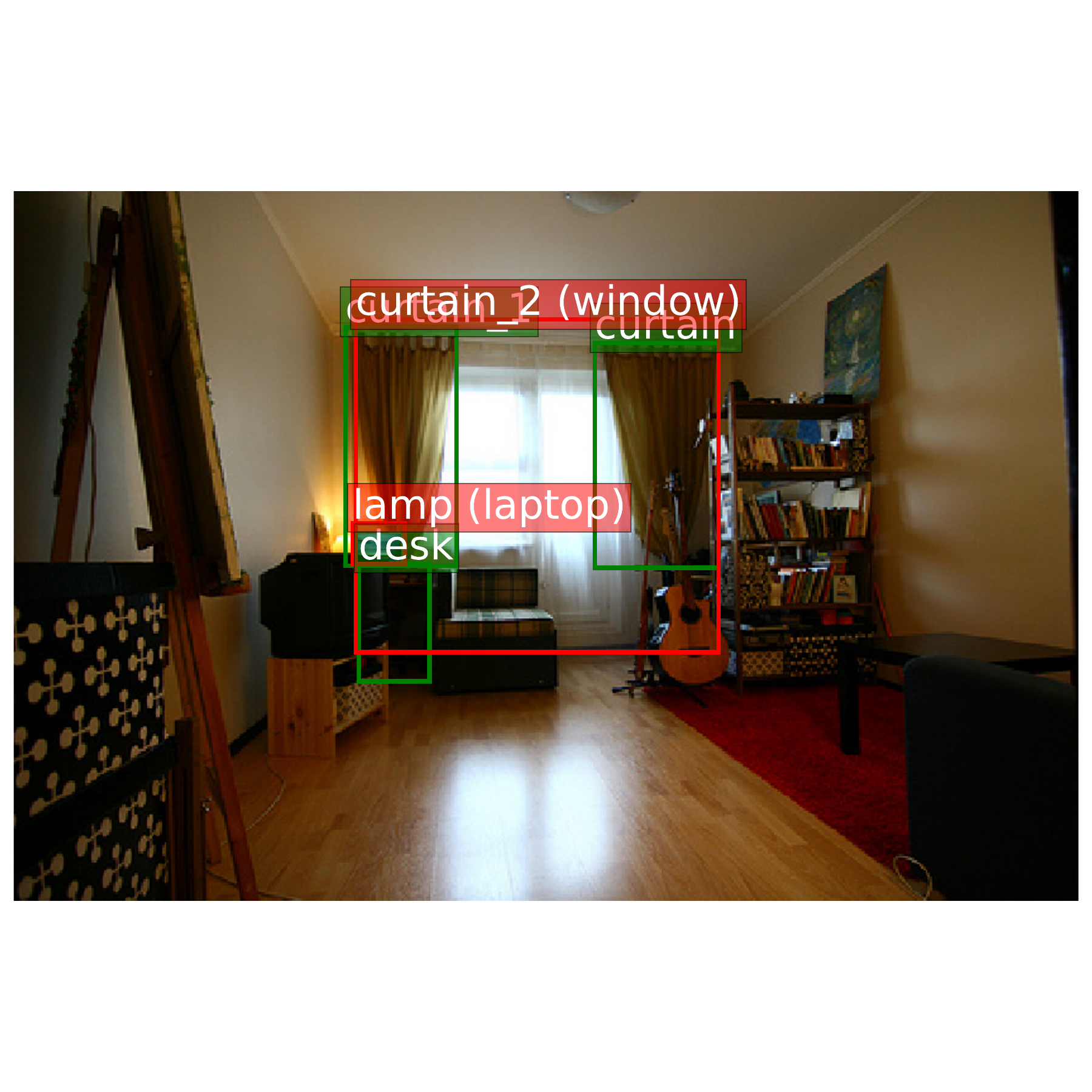}\\
\vspace{-1.1cm}
\includegraphics[width=.41\linewidth]{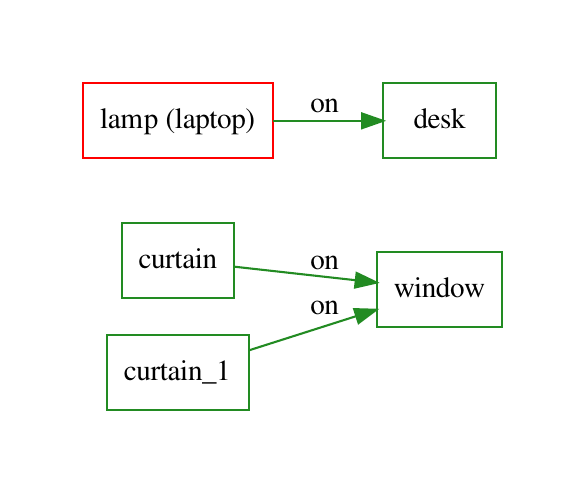}\hfill
\includegraphics[width=.53\linewidth]{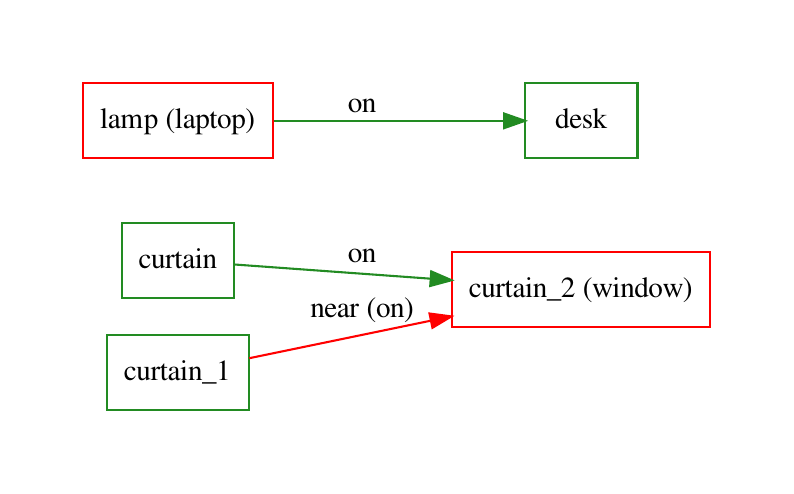}\hspace{-0.7cm}
\vspace{-1.1cm}
\end{center}
   \caption{Example comparison of our method \textsc{GB-Net} (left) with KERN \cite{chen2019knowledge} (right). Our method correctly detects the two pieces of curtain on window, while KERN predicts the less appropriate triplet curtain on curtain, possibly because the bounding box of the window contains the curtain as well.}
\label{fig:example_curtain}
\end{figure*}

\begin{figure*}
\begin{center}
\includegraphics[width=.45\linewidth]{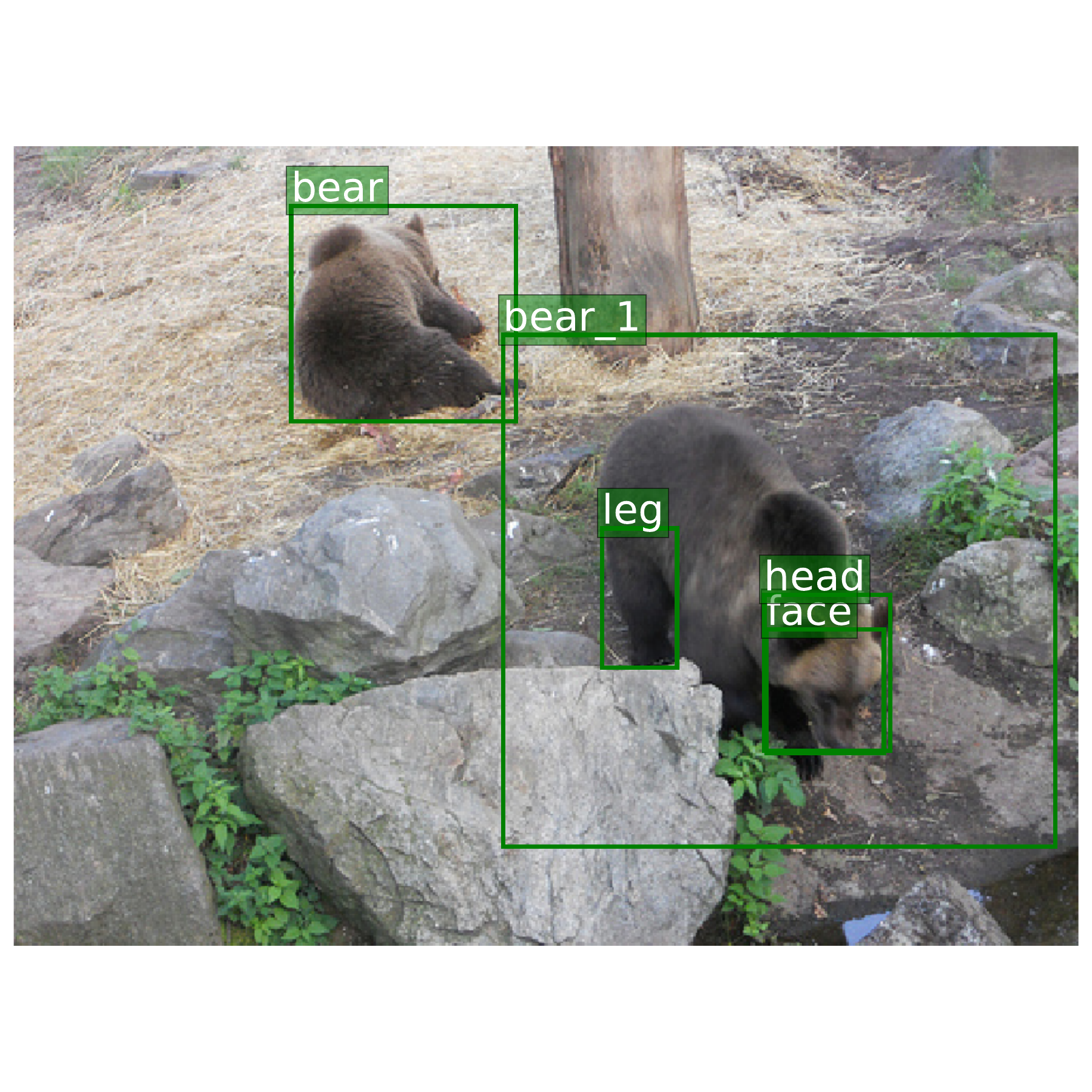}\hfill
\includegraphics[width=.45\linewidth]{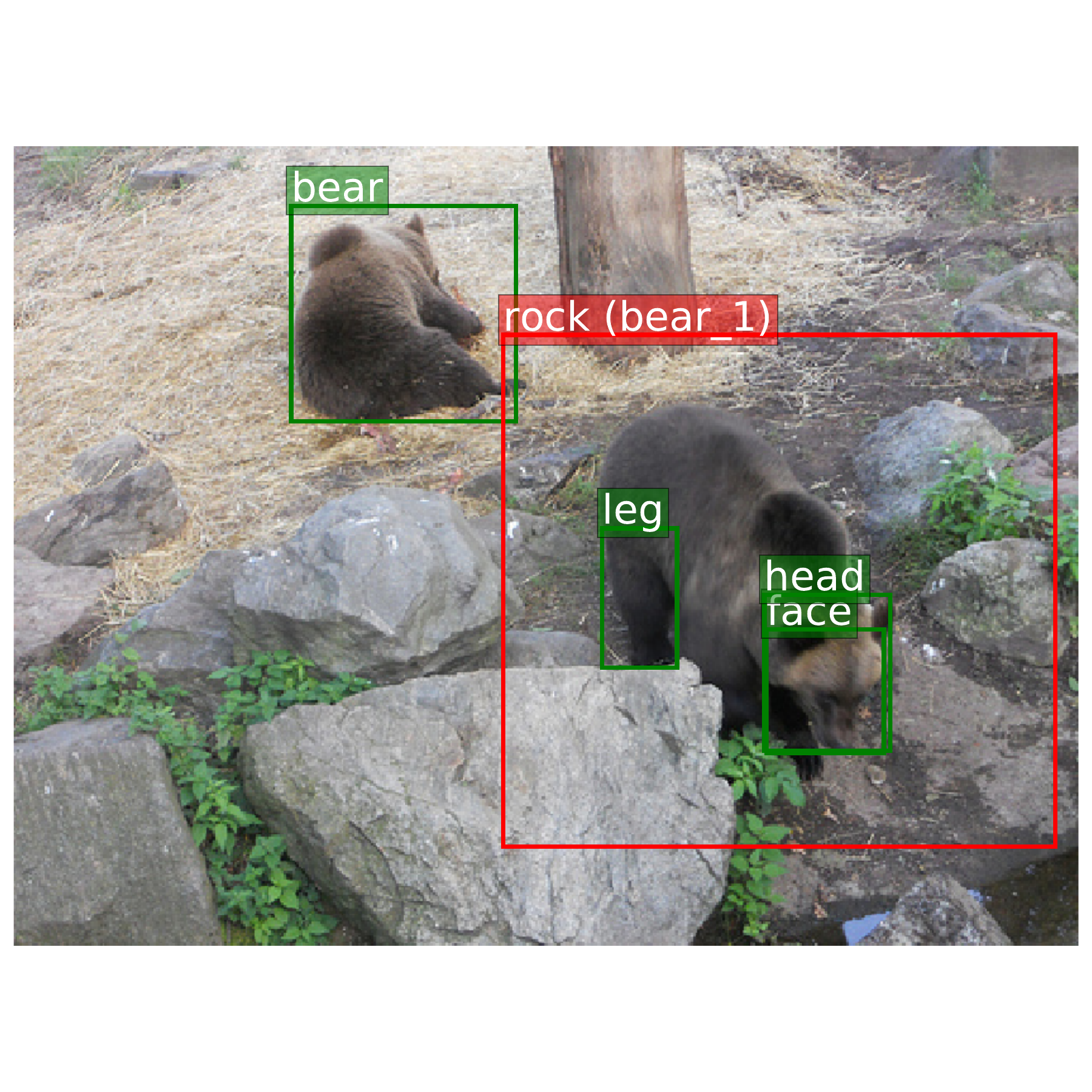}\\
\vspace{-1.0cm}

\scalebox{.75}{
\hspace{1cm}\includegraphics[width=.4\linewidth]{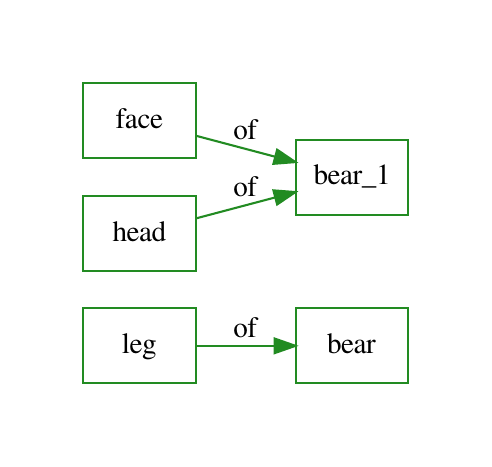}\hfill\hspace{2.5cm}
\includegraphics[width=.5\linewidth]{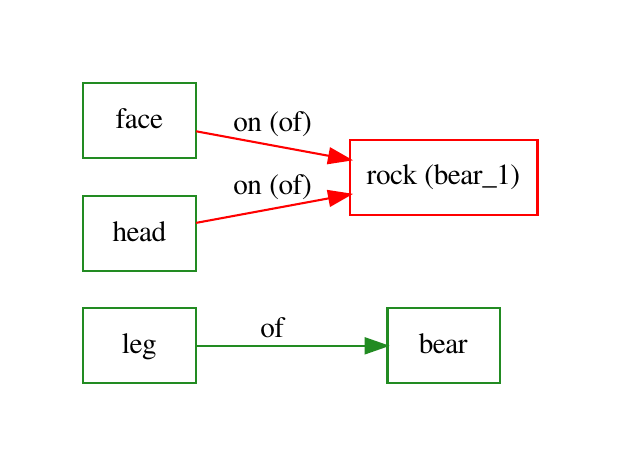}
}
\vspace{-0.9cm}
\end{center}
   \caption{Example comparison of our method \textsc{GB-Net} (left) with KERN \cite{chen2019knowledge} (right). KERN misclassifies the bear as rock, possibly due to the too loose bounding box that includes rocks as well. This leads to nonsensical triplets such as face on rock and head on rock, while our method produces more likely and accurate triplets face of bear and head of bear.}
\label{fig:example_bear}
\end{figure*}

\begin{figure*}
\begin{center}
\includegraphics[width=.45\linewidth]{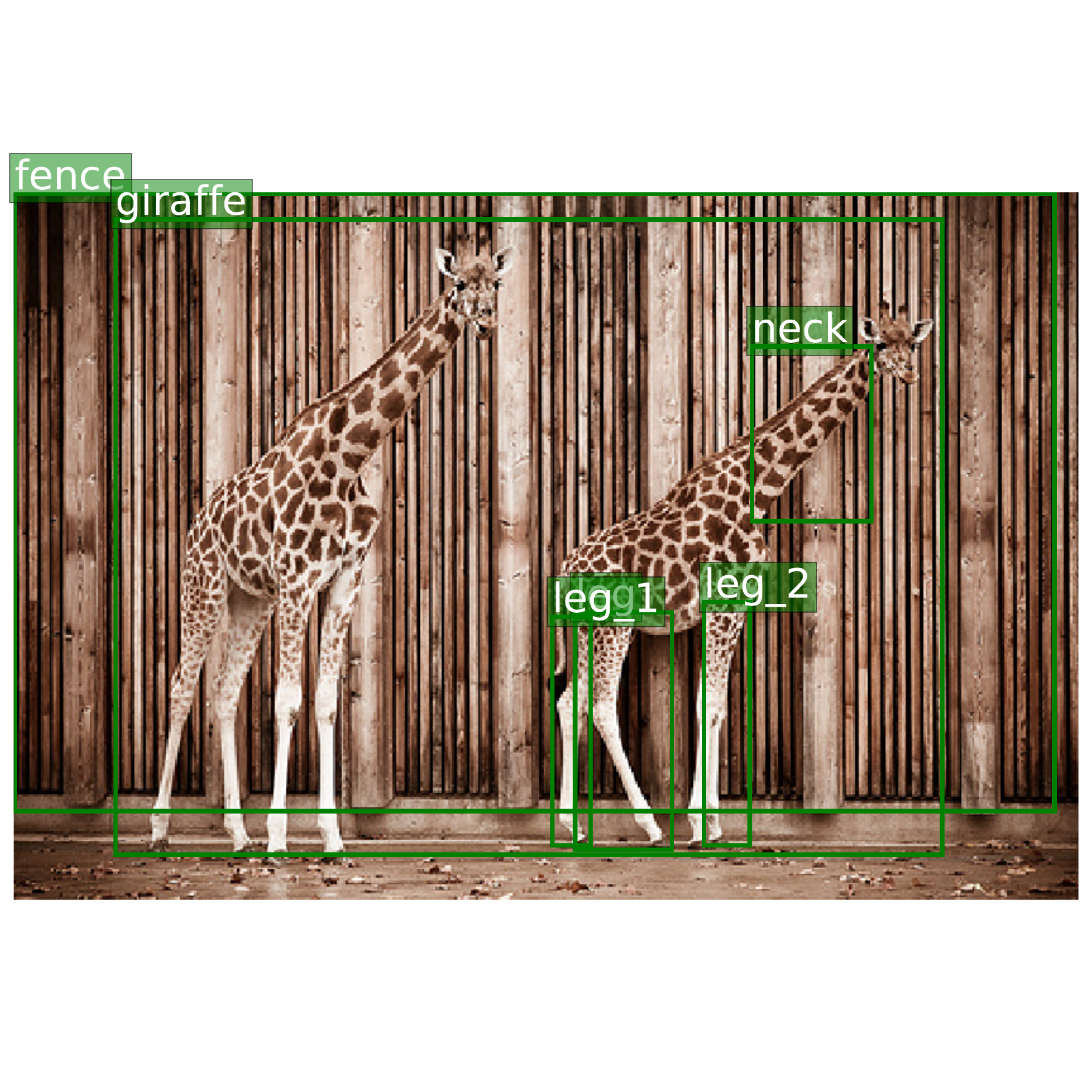}\hfill
\includegraphics[width=.45\linewidth]{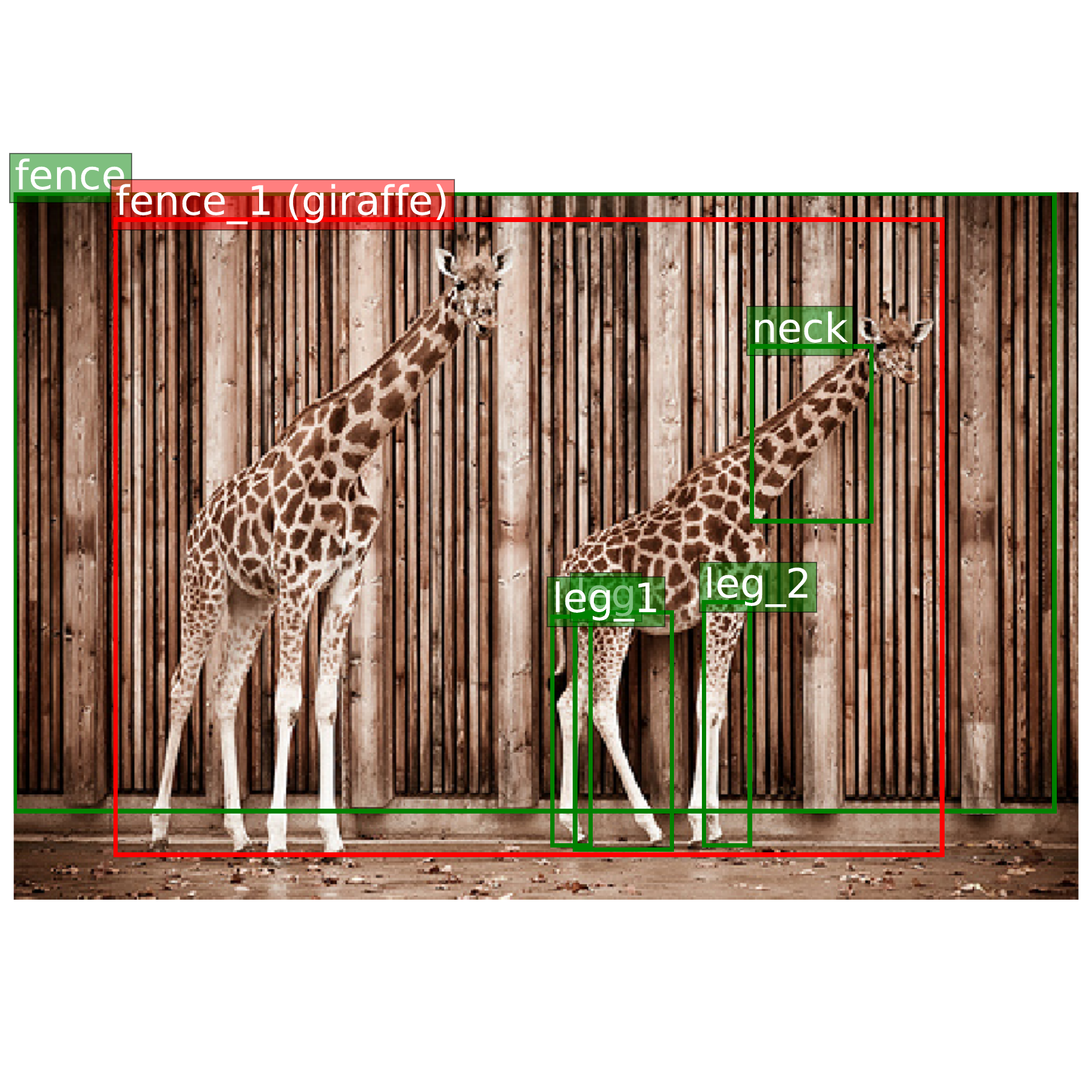}\\
\vspace{-1.0cm}
\includegraphics[width=.46\linewidth]{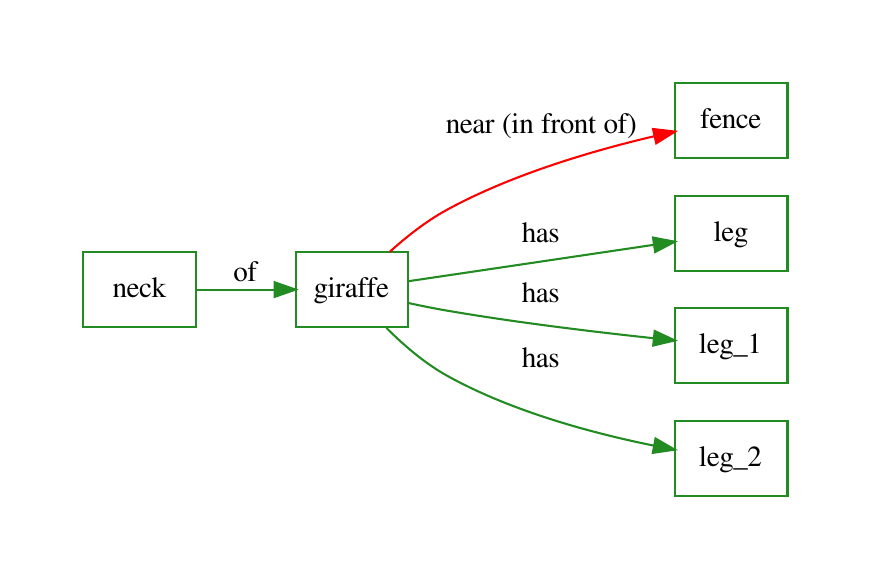}\hfill
\includegraphics[width=.53\linewidth]{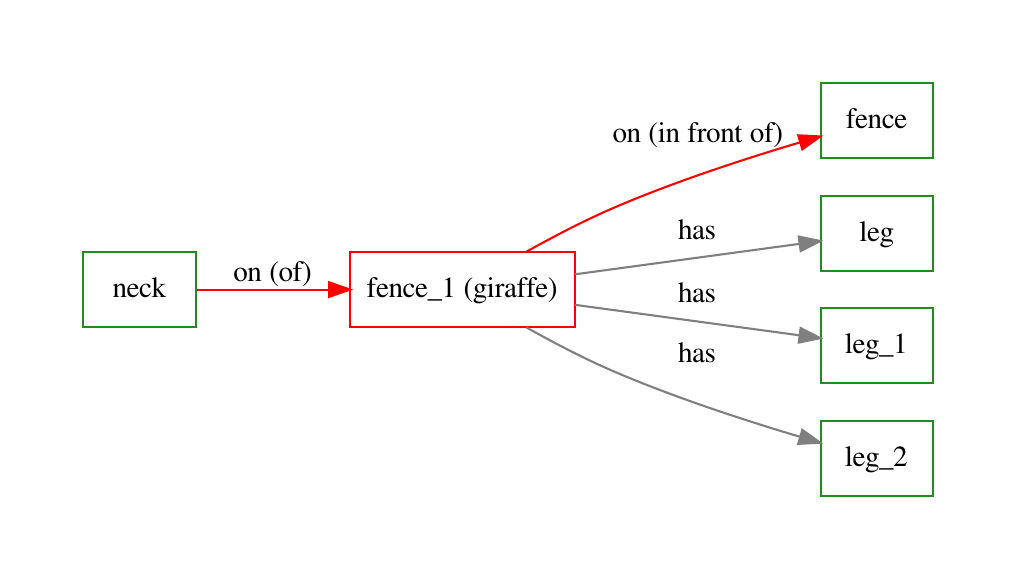}\hspace{-0.7cm}
\vspace{-0.8cm}
\end{center}
   \caption{Example comparison of our method \textsc{GB-Net} (left) with KERN \cite{chen2019knowledge} (right). KERN misclassifies the giraffe as a fence, leading to nonsensical triplets such as fence on fence, fence has leg, etc. Our method avoids such inappropriates compositions.}
\label{fig:example_giraffe}
\end{figure*}

\begin{figure*}
\begin{center}
\includegraphics[width=.45\linewidth]{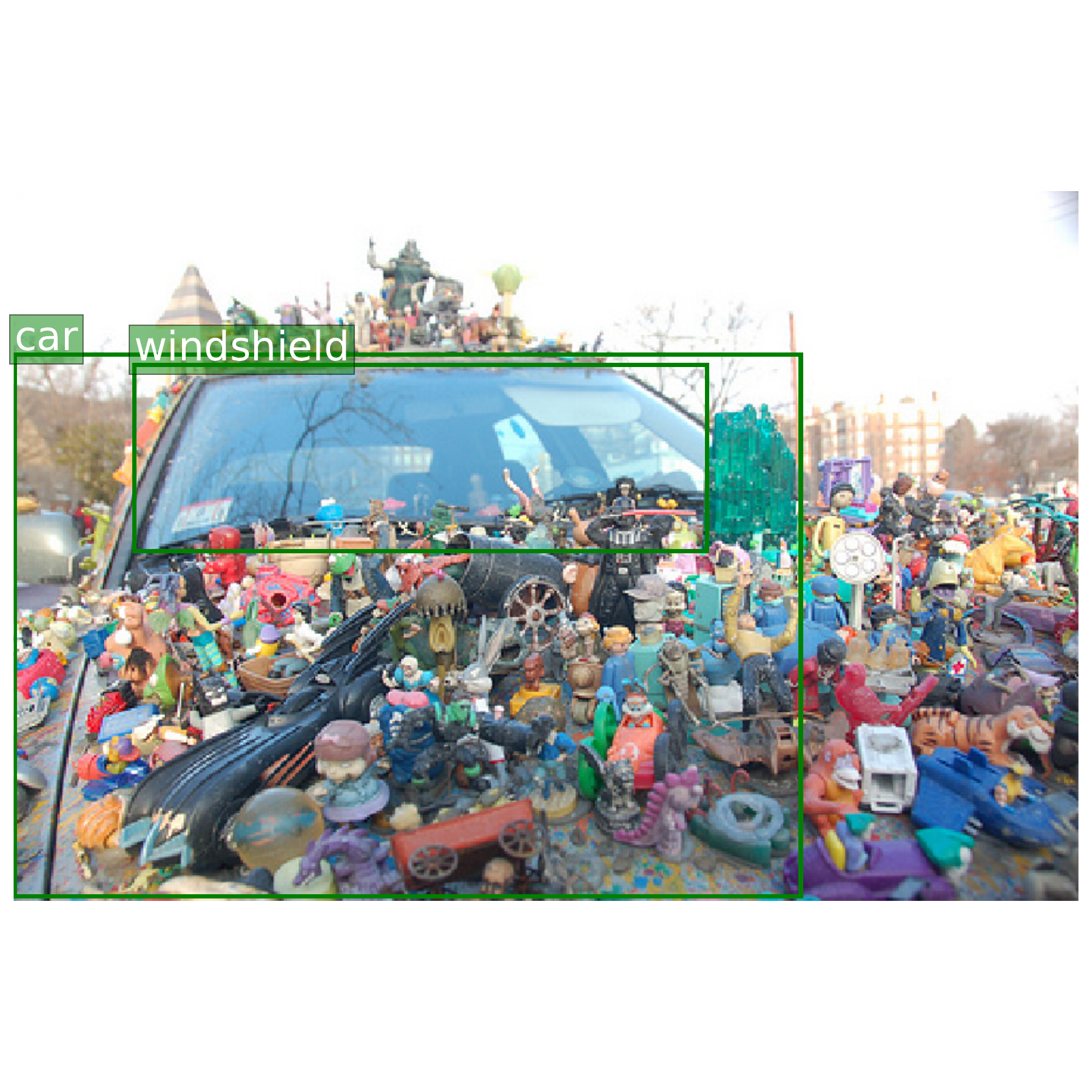}\hfill
\includegraphics[width=.45\linewidth]{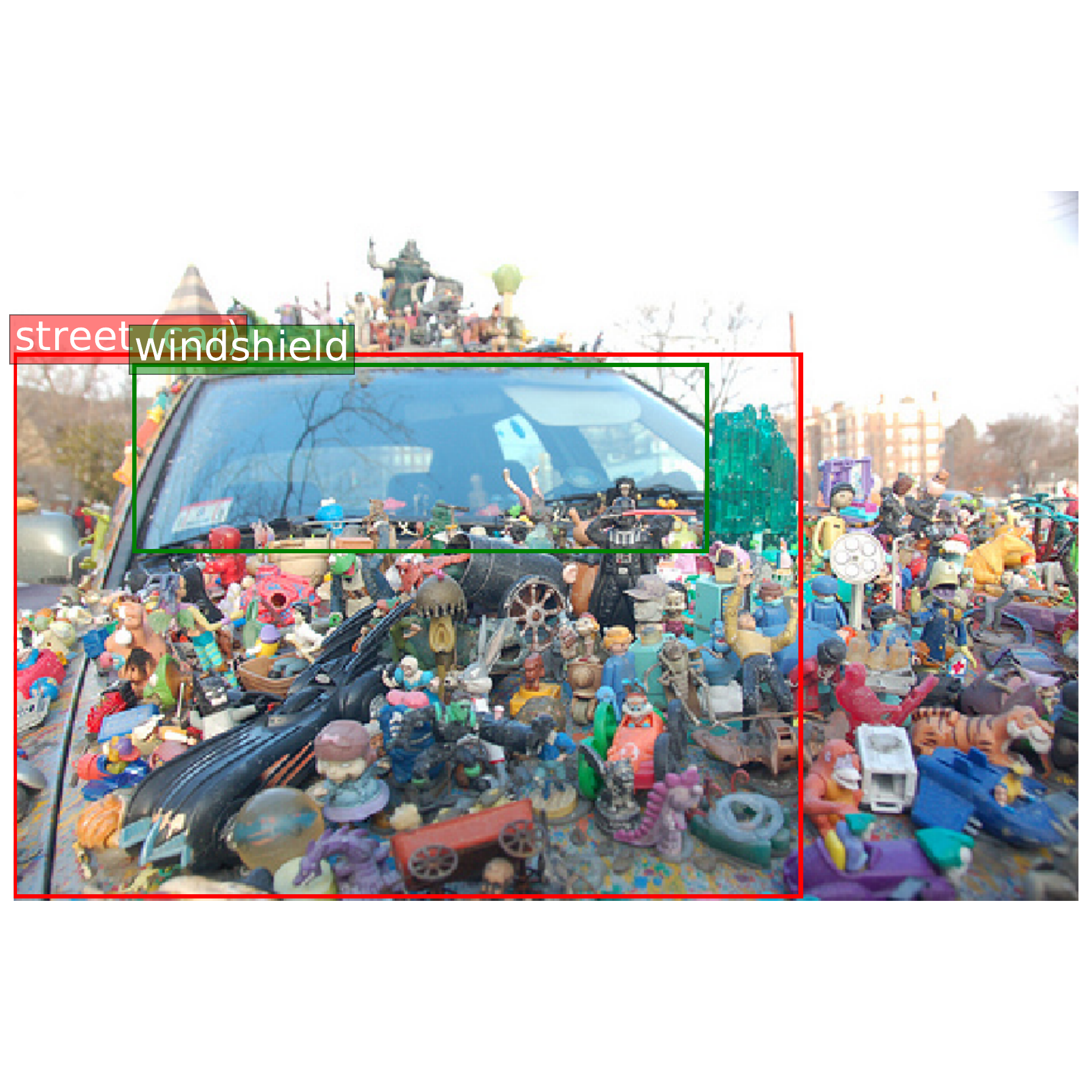}\\
\vspace{-1cm}
\scalebox{0.7}{
\includegraphics[width=.48\linewidth]{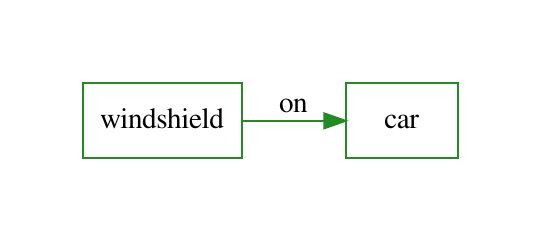}\hfill\hspace{2.5cm}
\includegraphics[width=.48\linewidth]{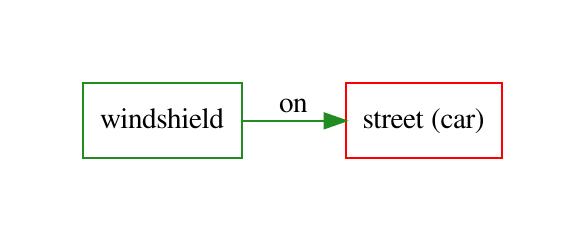}
}
\vspace{-0.8cm}
\end{center}
   \caption{Example comparison of our method \textsc{GB-Net} (left) with KERN \cite{chen2019knowledge} (right). KERN misclassifies car as street due to the extreme occlusion, while our method exploits the fact that cars are more likely to have windshields than streets.}
\label{fig:example_car_windshield}
\end{figure*}

%% file: supp/5_code.tex
\section{Software package\label{sec:code}}

We will provide a software package that reproduces every single reported number, \textit{i.e.}, all numbers in Table 1 and 2 of the main paper. 
To make it easy to reproduce, we provide an IPython Notebook for each experiment. We also provide a README file with a step by step guide, as well as a mapping between the notebook files and table cells in the paper. 
To reproduce the results from scratch, one would run the training notebook of each experiment followed by the evaluation notebook. 
To bypass training, we provide all the parameter checkpoints through a link in the README. This way the readers only need to run the evaluation notebook. 
In case a GPU is not available for deploying the model, we also provide a link to the pre-computed model outputs in the README. 
Finally, the notebooks already contain the saved evaluation results within them, which can be checked without running evaluation at all.

%% file: eccv2020submission.bbl
\begin{thebibliography}{10}
\providecommand{\url}[1]{\texttt{#1}}
\providecommand{\urlprefix}{URL }
\providecommand{\doi}[1]{https://doi.org/#1}

\bibitem{bosselut2019comet}
Bosselut, A., Rashkin, H., Sap, M., Malaviya, C., Celikyilmaz, A., Choi, Y.:
  Comet: Commonsense transformers for automatic knowledge graph construction.
  arXiv preprint arXiv:1906.05317  (2019)

\bibitem{chen2019knowledge}
Chen, T., Yu, W., Chen, R., Lin, L.: Knowledge-embedded routing network for
  scene graph generation. In: Proceedings of the IEEE Conference on Computer
  Vision and Pattern Recognition. pp. 6163--6171 (2019)

\bibitem{chen2018iterative}
Chen, X., Li, L.J., Fei-Fei, L., Gupta, A.: Iterative visual reasoning beyond
  convolutions. In: Proceedings of the IEEE Conference on Computer Vision and
  Pattern Recognition. pp. 7239--7248 (2018)

\bibitem{cho2014learning}
Cho, K., Van~Merri{\"e}nboer, B., Gulcehre, C., Bahdanau, D., Bougares, F.,
  Schwenk, H., Bengio, Y.: Learning phrase representations using rnn
  encoder-decoder for statistical machine translation. arXiv preprint
  arXiv:1406.1078  (2014)

\bibitem{cui2019class}
Cui, Y., Jia, M., Lin, T.Y., Song, Y., Belongie, S.: Class-balanced loss based
  on effective number of samples. In: Proceedings of the IEEE Conference on
  Computer Vision and Pattern Recognition. pp. 9268--9277 (2019)

\bibitem{dietz2018utilizing}
Dietz, L., Kotov, A., Meij, E.: Utilizing knowledge graphs for text-centric
  information retrieval. In: The 41st International ACM SIGIR Conference on
  Research \& Development in Information Retrieval. pp. 1387--1390. ACM (2018)

\bibitem{fader2014open}
Fader, A., Zettlemoyer, L., Etzioni, O.: Open question answering over curated
  and extracted knowledge bases. In: Proceedings of the 20th ACM SIGKDD
  international conference on Knowledge discovery and data mining. pp.
  1156--1165. ACM (2014)

\bibitem{flanigan2014discriminative}
Flanigan, J., Thomson, S., Carbonell, J., Dyer, C., Smith, N.A.: A
  discriminative graph-based parser for the abstract meaning representation.
  In: Proceedings of the 52nd Annual Meeting of the Association for
  Computational Linguistics (Volume 1: Long Papers). pp. 1426--1436 (2014)

\bibitem{gardner2018neural}
Gardner, M., Dasigi, P., Iyer, S., Suhr, A., Zettlemoyer, L.: Neural semantic
  parsing. In: Proceedings of the 56th Annual Meeting of the Association for
  Computational Linguistics: Tutorial Abstracts. pp. 17--18 (2018)

\bibitem{gu2019scene}
Gu, J., Zhao, H., Lin, Z., Li, S., Cai, J., Ling, M.: Scene graph generation
  with external knowledge and image reconstruction. In: Proceedings of the IEEE
  Conference on Computer Vision and Pattern Recognition. pp. 1969--1978 (2019)

\bibitem{herzig2018mapping}
Herzig, R., Raboh, M., Chechik, G., Berant, J., Globerson, A.: Mapping images
  to scene graphs with permutation-invariant structured prediction. In:
  Advances in Neural Information Processing Systems. pp. 7211--7221 (2018)

\bibitem{hudson2019learning}
Hudson, D.A., Manning, C.D.: Learning by abstraction: The neural state machine.
  arXiv preprint arXiv:1907.03950  (2019)

\bibitem{ilievski2020consolidating}
Ilievski, F., Szekely, P., Cheng, J., Zhang, F., Qasemi, E.: Consolidating
  commonsense knowledge. arXiv preprint arXiv:2006.06114  (2020)

\bibitem{johnson2018image}
Johnson, J., Gupta, A., Fei-Fei, L.: Image generation from scene graphs. In:
  Proceedings of the IEEE Conference on Computer Vision and Pattern
  Recognition. pp. 1219--1228 (2018)

\bibitem{johnson2015image}
Johnson, J., Krishna, R., Stark, M., Li, L.J., Shamma, D., Bernstein, M.,
  Fei-Fei, L.: Image retrieval using scene graphs. In: Proceedings of the IEEE
  conference on computer vision and pattern recognition. pp. 3668--3678 (2015)

\bibitem{kato2018compositional}
Kato, K., Li, Y., Gupta, A.: Compositional learning for human object
  interaction. In: Proceedings of the European Conference on Computer Vision
  (ECCV). pp. 234--251 (2018)

\bibitem{khashabi2018question}
Khashabi, D., Khot, T., Sabharwal, A., Roth, D.: Question answering as global
  reasoning over semantic abstractions. In: Thirty-Second AAAI Conference on
  Artificial Intelligence (2018)

\bibitem{kingma2014adam}
Kingma, D.P., Ba, J.: Adam: A method for stochastic optimization. arXiv
  preprint arXiv:1412.6980  (2014)

\bibitem{kipf2016semi}
Kipf, T.N., Welling, M.: Semi-supervised classification with graph
  convolutional networks. arXiv preprint arXiv:1609.02907  (2016)

\bibitem{krishna2017visual}
Krishna, R., Zhu, Y., Groth, O., Johnson, J., Hata, K., Kravitz, J., Chen, S.,
  Kalantidis, Y., Li, L.J., Shamma, D.A., et~al.: Visual genome: Connecting
  language and vision using crowdsourced dense image annotations. International
  Journal of Computer Vision  \textbf{123}(1),  32--73 (2017)

\bibitem{lee2018multi}
Lee, C.W., Fang, W., Yeh, C.K., Frank~Wang, Y.C.: Multi-label zero-shot
  learning with structured knowledge graphs. In: Proceedings of the IEEE
  Conference on Computer Vision and Pattern Recognition. pp. 1576--1585 (2018)

\bibitem{li2019multilingual}
Li, M., Lin, Y., Hoover, J., Whitehead, S., Voss, C., Dehghani, M., Ji, H.:
  Multilingual entity, relation, event and human value extraction. In:
  Proceedings of the 2019 Conference of the North American Chapter of the
  Association for Computational Linguistics (Demonstrations). pp. 110--115
  (2019)

\bibitem{li2018factorizable}
Li, Y., Ouyang, W., Zhou, B., Shi, J., Zhang, C., Wang, X.: Factorizable net:
  an efficient subgraph-based framework for scene graph generation. In:
  Proceedings of the European Conference on Computer Vision (ECCV). pp.
  335--351 (2018)

\bibitem{li2017scene}
Li, Y., Ouyang, W., Zhou, B., Wang, K., Wang, X.: Scene graph generation from
  objects, phrases and region captions. In: Proceedings of the IEEE
  International Conference on Computer Vision. pp. 1261--1270 (2017)

\bibitem{li2015gated}
Li, Y., Tarlow, D., Brockschmidt, M., Zemel, R.: Gated graph sequence neural
  networks. arXiv preprint arXiv:1511.05493  (2015)

\bibitem{liang2018visual}
Liang, K., Guo, Y., Chang, H., Chen, X.: Visual relationship detection with
  deep structural ranking. In: Thirty-Second AAAI Conference on Artificial
  Intelligence (2018)

\bibitem{liu2004conceptnet}
Liu, H., Singh, P.: Conceptnet—a practical commonsense reasoning tool-kit. BT
  technology journal  \textbf{22}(4),  211--226 (2004)

\bibitem{liu2018structure}
Liu, Y., Wang, R., Shan, S., Chen, X.: Structure inference net: Object
  detection using scene-level context and instance-level relationships. In:
  Proceedings of the IEEE conference on computer vision and pattern
  recognition. pp. 6985--6994 (2018)

\bibitem{marino2016more}
Marino, K., Salakhutdinov, R., Gupta, A.: The more you know: Using knowledge
  graphs for image classification. arXiv preprint arXiv:1612.04844  (2016)

\bibitem{miller1995wordnet}
Miller, G.A.: Wordnet: a lexical database for english. Communications of the
  ACM  \textbf{38}(11),  39--41 (1995)

\bibitem{newell2017pixels}
Newell, A., Deng, J.: Pixels to graphs by associative embedding. In: Advances
  in neural information processing systems. pp. 2171--2180 (2017)

\bibitem{pei2011parsing}
Pei, M., Jia, Y., Zhu, S.C.: Parsing video events with goal inference and
  intent prediction. In: 2011 International Conference on Computer Vision. pp.
  487--494. IEEE (2011)

\bibitem{pennington2014glove}
Pennington, J., Socher, R., Manning, C.: Glove: Global vectors for word
  representation. In: Proceedings of the 2014 conference on empirical methods
  in natural language processing (EMNLP). pp. 1532--1543 (2014)

\bibitem{qi2019attentive}
Qi, M., Li, W., Yang, Z., Wang, Y., Luo, J.: Attentive relational networks for
  mapping images to scene graphs. In: Proceedings of the IEEE Conference on
  Computer Vision and Pattern Recognition. pp. 3957--3966 (2019)

\bibitem{qi2018learning}
Qi, S., Wang, W., Jia, B., Shen, J., Zhu, S.C.: Learning human-object
  interactions by graph parsing neural networks. In: Proceedings of the
  European Conference on Computer Vision (ECCV). pp. 401--417 (2018)

\bibitem{ren2015faster}
Ren, S., He, K., Girshick, R., Sun, J.: Faster r-cnn: Towards real-time object
  detection with region proposal networks. In: Advances in neural information
  processing systems. pp. 91--99 (2015)

\bibitem{schuster2015generating}
Schuster, S., Krishna, R., Chang, A., Fei-Fei, L., Manning, C.D.: Generating
  semantically precise scene graphs from textual descriptions for improved
  image retrieval. In: Proceedings of the fourth workshop on vision and
  language. pp. 70--80 (2015)

\bibitem{shi2019explainable}
Shi, J., Zhang, H., Li, J.: Explainable and explicit visual reasoning over
  scene graphs. In: Proceedings of the IEEE Conference on Computer Vision and
  Pattern Recognition. pp. 8376--8384 (2019)

\bibitem{teney2017graph}
Teney, D., Liu, L., van~den Hengel, A.: Graph-structured representations for
  visual question answering. In: Proceedings of the IEEE Conference on Computer
  Vision and Pattern Recognition. pp.~1--9 (2017)

\bibitem{tu2014joint}
Tu, K., Meng, M., Lee, M.W., Choe, T.E., Zhu, S.C.: Joint video and text
  parsing for understanding events and answering queries. IEEE MultiMedia
  \textbf{21}(2),  42--70 (2014)

\bibitem{wadden2019entity}
Wadden, D., Wennberg, U., Luan, Y., Hajishirzi, H.: Entity, relation, and event
  extraction with contextualized span representations. arXiv preprint
  arXiv:1909.03546  (2019)

\bibitem{wang2018zero}
Wang, X., Ye, Y., Gupta, A.: Zero-shot recognition via semantic embeddings and
  knowledge graphs. In: Proceedings of the IEEE Conference on Computer Vision
  and Pattern Recognition. pp. 6857--6866 (2018)

\bibitem{woo2018linknet}
Woo, S., Kim, D., Cho, D., Kweon, I.S.: Linknet: Relational embedding for scene
  graph. In: Advances in Neural Information Processing Systems. pp. 560--570
  (2018)

\bibitem{xu2017scene}
Xu, D., Zhu, Y., Choy, C.B., Fei-Fei, L.: Scene graph generation by iterative
  message passing. In: Proceedings of the IEEE Conference on Computer Vision
  and Pattern Recognition. pp. 5410--5419 (2017)

\bibitem{xu2019spatial}
Xu, H., Jiang, C., Liang, X., Li, Z.: Spatial-aware graph relation network for
  large-scale object detection. In: Proceedings of the IEEE Conference on
  Computer Vision and Pattern Recognition. pp. 9298--9307 (2019)

\bibitem{xu2019reasoning}
Xu, H., Jiang, C., Liang, X., Lin, L., Li, Z.: Reasoning-rcnn: Unifying
  adaptive global reasoning into large-scale object detection. In: Proceedings
  of the IEEE Conference on Computer Vision and Pattern Recognition. pp.
  6419--6428 (2019)

\bibitem{yang2018graph}
Yang, J., Lu, J., Lee, S., Batra, D., Parikh, D.: Graph r-cnn for scene graph
  generation. In: Proceedings of the European Conference on Computer Vision
  (ECCV). pp. 670--685 (2018)

\bibitem{yang2019auto}
Yang, X., Tang, K., Zhang, H., Cai, J.: Auto-encoding scene graphs for image
  captioning. In: Proceedings of the IEEE Conference on Computer Vision and
  Pattern Recognition. pp. 10685--10694 (2019)

\bibitem{yao2018exploring}
Yao, T., Pan, Y., Li, Y., Mei, T.: Exploring visual relationship for image
  captioning. In: Proceedings of the European Conference on Computer Vision
  (ECCV). pp. 684--699 (2018)

\bibitem{yu2018modeling}
Yu, J., Lu, Y., Qin, Z., Zhang, W., Liu, Y., Tan, J., Guo, L.: Modeling text
  with graph convolutional network for cross-modal information retrieval. In:
  Pacific Rim Conference on Multimedia. pp. 223--234. Springer (2018)

\bibitem{zareian2020weakly}
Zareian, A., Karaman, S., Chang, S.F.: Weakly supervised visual semantic
  parsing. In: Proceedings of the IEEE/CVF Conference on Computer Vision and
  Pattern Recognition. pp. 3736--3745 (2020)

\bibitem{zellers2018neural}
Zellers, R., Yatskar, M., Thomson, S., Choi, Y.: Neural motifs: Scene graph
  parsing with global context. In: Proceedings of the IEEE Conference on
  Computer Vision and Pattern Recognition. pp. 5831--5840 (2018)

\bibitem{zhan2019exploring}
Zhan, Y., Yu, J., Yu, T., Tao, D.: On exploring undetermined relationships for
  visual relationship detection. In: Proceedings of the IEEE Conference on
  Computer Vision and Pattern Recognition. pp. 5128--5137 (2019)

\bibitem{zhang2019empirical}
Zhang, C., Chao, W.L., Xuan, D.: An empirical study on leveraging scene graphs
  for visual question answering. arXiv preprint arXiv:1907.12133  (2019)

\bibitem{zhao2011image}
Zhao, Y., Zhu, S.C.: Image parsing with stochastic scene grammar. In: Advances
  in Neural Information Processing Systems. pp. 73--81 (2011)

\end{thebibliography}
